\documentclass[journal, onecolumn]{IEEEtran}
\ifCLASSINFOpdf
\else
\fi

\usepackage{graphicx}
\usepackage{cite}
\usepackage{picinpar}
\usepackage{amsmath}
\usepackage{url}
\usepackage{colortbl}
\usepackage{soul}
\usepackage{multirow}
\usepackage{pifont}
\usepackage{color}
\usepackage{alltt}
\usepackage[hidelinks]{hyperref}
\usepackage{enumerate}
\usepackage{siunitx}
\usepackage{breakurl}
\usepackage{epstopdf}
\usepackage{pbox}

\usepackage{array}
\usepackage{threeparttable}
\usepackage{multirow}
\usepackage{algorithm}
\usepackage{algorithmic}
\usepackage{bm}
\usepackage{CJK}
\usepackage{indentfirst}
\usepackage{amsmath}
\usepackage{subfigure}
\usepackage{color}
\usepackage{booktabs}
\usepackage{amsmath,amssymb}
\usepackage{epstopdf}
\usepackage{textcomp}
\usepackage{diagbox}
\usepackage{setspace}

\usepackage{bm}
\usepackage{CJK}
\usepackage{indentfirst}
\usepackage{amsmath}
\usepackage{multirow}
\usepackage{threeparttable}
\usepackage{graphicx}
\usepackage{subfigure}
\usepackage{color}
\usepackage{booktabs}
\usepackage{epstopdf}

\DeclareMathAlphabet{\mathcal}{OMS}{cmsy}{m}{n}

\begin{document}
\title{A Transferable Multi-stage Model  with \\ Cycling Discrepancy Learning for \\ Lithium-ion Battery State of Health Estimation}

\author{Yan~Qin,~\IEEEmembership{Member,~IEEE,}
       	       Chau~Yuen,~\IEEEmembership{Fellow,~IEEE,}
	       	      Xunyuan~Yin,~\IEEEmembership{Member,~IEEE,}
        	  	Biao~Huang,~\IEEEmembership{Fellow,~IEEE}	
\thanks{Y. Qin and C. Yuen are with the Engineering Product Development Pillar, The Singapore University of Technology and Design, 8 Somapah Road, 487372 Singapore. (e-mail: zdqinyan@gmail.com, yuenchau@sutd.edu.sg) (Corresponding author: Chau Yuen)}
\thanks{X. Yin is with the School of Chemical and Biomedical Engineering, Nanyang Technological University, 637459, Singapore. (email: Xunyuan.Yin@ntu.edu.sg).}
\thanks{B. Huang is with the Department of Chemical and Materials Engineering, University of Alberta, Edmonton, Alberta, Canada. (e-mail: biao.huang@ualberta.ca)}
}

\maketitle
\begin{abstract}
As a significant ingredient regarding health status, data-driven state-of-health (SOH) estimation has become dominant for lithium-ion batteries (LiBs). To handle data discrepancy across batteries, current SOH estimation models engage in transfer learning (TL), which reserves apriori knowledge gained through reusing partial structures of the offline trained model. However, multiple degradation patterns of a complete life cycle of a battery make it challenging to pursue TL. The concept of the stage is introduced to describe the collection of continuous cycles that present a similar degradation pattern. A transferable multi-stage SOH estimation model is proposed to perform TL across batteries in the same stage, consisting of four steps. First, with identified stage information, raw cycling data from the source battery are reconstructed into the phase space with high dimensions, exploring hidden dynamics with limited sensors. Next, domain invariant representation across cycles in each stage is proposed through cycling discrepancy subspace with reconstructed data. Third, considering the unbalanced discharge cycles among different stages, a switching estimation strategy composed of a lightweight model with the long short-term memory network and a powerful model with the proposed temporal capsule network is proposed to boost estimation accuracy. Lastly, an updating scheme compensates for estimation errors when the cycling consistency of target batteries drifts. The proposed method outperforms its competitive algorithms in various transfer tasks for a run-to-failure benchmark with three batteries. Especially through transferring the estimation model from batteries B7 to B6, the proposed method improves the estimation accuracy by as high as $42.6\%$ in the third stage in terms of the root mean square error, compared to other state-of-the-art approaches. In addition, similar conclusions can be drawn from other contributed experiments.
\end{abstract}

\begin{IEEEkeywords}
Lithium-ion battery, transfer learning, multi-stage performance degradation, state of health estimation.
\end{IEEEkeywords}
\IEEEpeerreviewmaketitle

\section{Introduction}
\IEEEPARstart{T}{he} ever-challenging environmental concerns and global warming crisis necessitate sustainable energy as an alternative to traditional fossil fuels, e.g., oil and coal. The emergence of Lithium-ion batteries (LiBs) carries an elevated influence on the current energy system. With the advancements in material science and manufacturing technology, rechargeable LiBs have become an eco-friendly choice in our daily and commercial electronics devices \cite{2018BatteryReview}, \cite{2019BatteryNature}. Although LiBs have become increasingly powerful, their performance is still subject to gradual degradation over time due to electrochemical side reactions in anode, electrolyte, and cathode \cite{BIRKL2017}. To provide a safe working environment for LiBs and their powered devices, the state of health (SOH) estimation plays a vital role in evaluating health status.


Nowadays, substantial work relies solely on historical degradation data to infer accurate SOH by imitating domain experts and constructing artificial intelligence systems with advanced machine learning approaches. Besides the remarkable breakthroughs in distributed computation and communication technology, algorithmic efficacy and data availability have become the main concerns in pursuing accurate SOH estimation for both training and testing data. Generally speaking, for typical estimation modelling, the online testing data are expected to follow a distribution or feature space with consistent offline training data \cite{2021Editor3}.

However, the prerequisite mentioned above does not always hold when the cycling data of LiBs are involved. This could be ascribed to two common factors given as follows. First, the differences in the manufacturing phase, such as material formulation, production technology, battery size, etc., will lead to LiBs used in training being different from others used in testing physically \cite{2021TIISOC_QIN}. Second, in the usage phase, a battery cell suffers from performance degradation irregularly over cycles caused by inherent loss of anode and cathode due to various operating conditions, leading to a non-consistent degradation pattern. In practice, it is time-consuming and high labor cost to collect sufficient run-to-failure cycling data under various operating conditions. These challenges make current SOH estimation models suffer from severe data discrepancy from two aspects, one of which is the instinct cycles in the same battery cell and the other one is the cycles from different LiBs.


By leveraging apriori knowledge learned from a source domain to a different but related target domain \cite{2010TKDE_Transferlearning}, transfer learning (TL) shows a promising solution to address the aforementioned  issues. For rotating machinery diagnosis, Han et al. \cite{2021Editor3} proposed a domain generalization-based model to explore the robust features from unseen conditions, and remarkable results are gained with the application to gearbox fault classification. Herein, we exemplify what may happen for SOH estimation using TL. The data and corresponding distribution during the training and testing phases are referred to as source domain and target domain in the TL framework, respectively. During TL, the learned mapping relationship from source data allows for accurate predictions in the target domain even though they have different data distributions.

Recently, with the development of advanced machine learning models, extensive research has focused on designing flexible networks to gain transferable mapping functions. Attracted by the promising ability in long sequence prediction, the long short-term memory (LSTM) network has been widely adopted as the favored network when massive labeled samples are available. Alternatively, Qian et al. \cite{2021RUL} put forward a novel hybrid deep learning network to alleviate the heavy reliance on labeled samples for supervised learning with the help of positive-unlabeled learning. With the combination of LSTM and fully connected layers, Tan et al. \cite{2020TIETrasnferLSTM} gained model transferability by freezing LSTM layers trained with source data. Then the authors used the target data to update fully connected layers, capturing specific characteristics of the target task. In the event of a small amount of source data, Shen et al. \cite{SHEN2020114296} introduced an ensemble learning mechanism to improve model robustness. The methods mentioned above belong to the parameter transfer-based fine-tuning \cite{2021Editor3}, which gains transferability by directly applying the well-trained offline network to the target task. Network fine-tuning-based TL is appealing due to the easy implementation without feature engineering but is limited by the lacking of interpretability. As an effective way to achieve the dual purposes of transferability and interpretability, domain invariant representation has been extensively studied to achieve transferable estimation purposes. By projecting original data into the shared space, the derived transfer components across the source and target domains are expected to preserve the common prediction ability \cite{2011TNNLS}. Ma et al. \cite{MA2021116167} designed a transferable LSTM using domain invariant samples to fit the degradation behavior of the target battery. Che et al. \cite{2021TVT_SOHtrransfer} employed a Euclidean distance-based index to select similar health indicators between source and target tasks. Similar indicators are used to initialize the SOH estimation model aimed at the target task. However, filtering out specific samples or constructing invariant domain indicators is high expert knowledge-extensive.

The aforementioned methods gain remarkable successes and pave the way to TL. It is worth mentioning that several limitations still need to be further addressed when a transferable SOH estimation model is targeted:
\begin{itemize}
\item Domain invariant representation with consistent predictive capability between the source and target domains has not received enough attention for battery SOH estimation.
\item Current SOH estimation models assume to follow a single degradation manner of the complete life cycle, which may not reflect the electrochemical physics.
\end{itemize}

These limitations mentioned-above are primarily attributed to the complex LiB data structure, which are inconsistent cycle-varying data. Precisely, a sample of degraded LiB composes of a two-dimension data matrix with variable data characteristics caused by the performance fading rather than a vector at a specific time. The two-dimensional data distribution varies over cycles, even in the training dataset. Meanwhile, degradation behavior may experience multiple patterns over cycles. That is, the degradation pattern maintains consistency within the same stage but is dissimilar between different stages. Here, the stage is a collection of continuous cycles to present a similar degradation pattern. Therefore, transferring SOH estimation ability in each stage is meaningful but not solved yet. To overcome these limitations, we argue that by capturing discrepancy and similarity over cycles for the same LiB or different LiBs and by extracting relevant parts from measurements regarding performance fading can contribute to further boosting estimation accuracy and interpretability. It makes sense to identify consistent features in each cycle that are similar over cycles through the interpretable embedding, uncovering inconsistent features relevant to performance fading to perform domain-invariant representation. Two critical issues need to be solved. The first one is to decompose the original cycling data with consistent and inconsistent behavior in an unsupervised learning way. Second, with the derived inconsistent components over training data, how to achieve transferable estimation with domain invariant representation.

In light of the aforementioned discussions, this article targets a transferable multi-stage SOH estimation model, which involves two parts. The first part focuses on domain invariant feature extraction, which solves the first challenge to distinguish the consistent components from inconsistent components over cycles. To reveal unseen dynamics from discharging measurements, raw cycling data are reconstructed in the phase space with the aid of known stage division information. In the reconstructed space, the subtle difference over cycles can be amplified. Next, the cycling consistency subspace and cycling discrepancy subspace are iteratively located by maximizing the cycle consistency calculated by Kullback-Leibler divergence in the same stage. The cycling discrepancy subspace finds the domain invariant representation responsible for predicting different LiBs and clearly explains the degradation behavior in a data-driven way. In the second part, the amounts of cycling data in each stage are unbalanced, requiring a flexible estimation framework by combining LSTM and the proposed temporal capsule network (TemCap). Having the temporal learning ability, TemCap possesses the high-dimension feature representation and avoids information loss compared to the convolutional neural network. Finally, with the developed offline model, online application achieves transferability by taking cycling discrepancy components over cycles from the source task as input, overcoming varying distribution caused by battery discrepancy and multiple degradation patterns. Updating strategy is triggered when the cycling consistency of the target domain deviates from source models. Otherwise, source models can be directly adopted for estimation. A series of comparisons verify the efficacy of the proposed method compared to its counterparts on a benchmark. The contributions of this work are summarized as follows:

\begin{figure}[!ht]
\centering
\begin{minipage}[t]{1\linewidth}
\centering
\includegraphics[width=9cm]{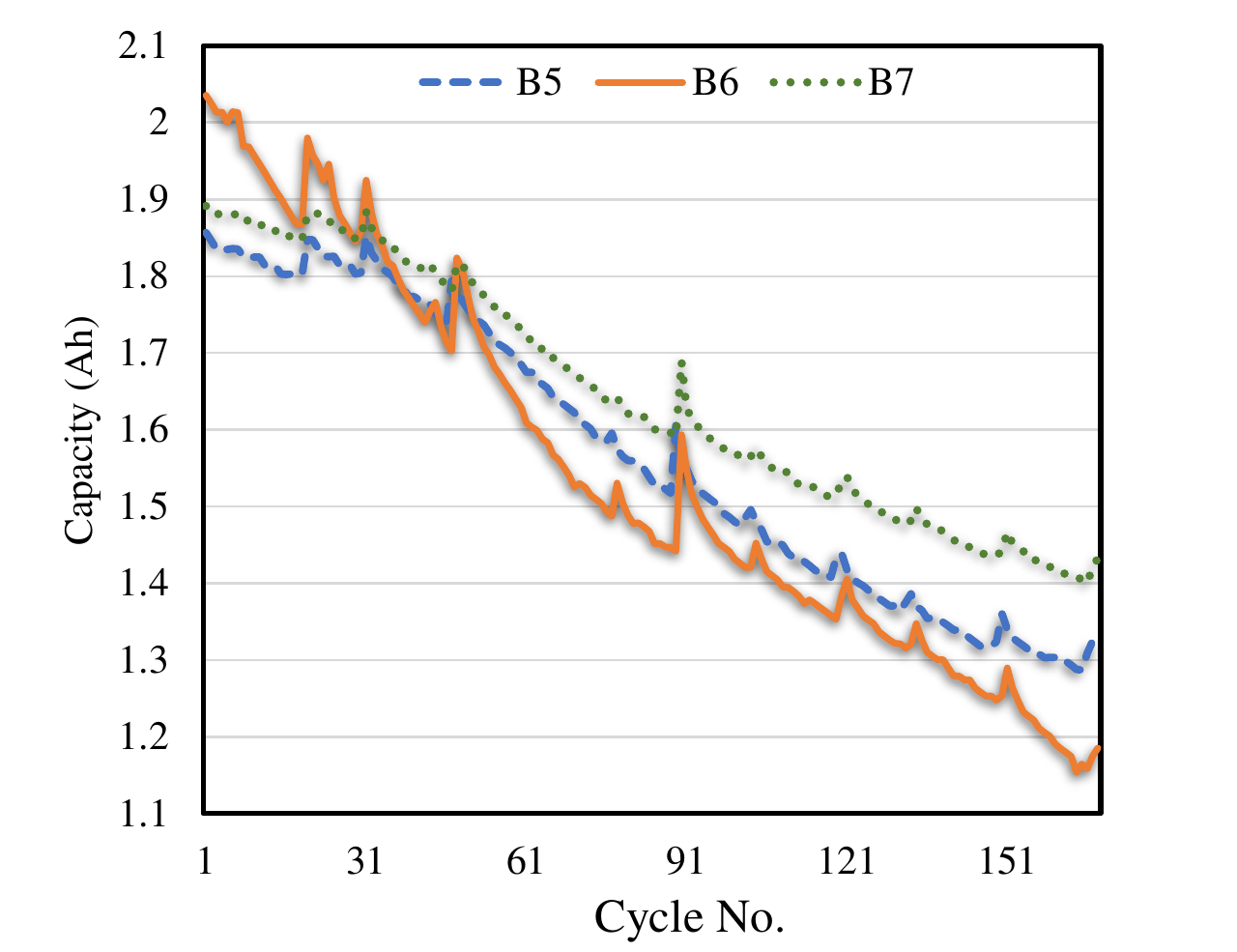}
\end{minipage}
\caption{Visualization of capacity fading procedure with multi-stage characteristics of Batteries B5, B6, and B7 in the NASA dataset \cite{2007NASA}.}
\label{FIG1}
\end{figure}

(1) Raw cycling data are reconstructed in the phase space, exploring unseen dynamics with limited sensors.

(2) Domain invariant representation is described through cycling discrepancy subspace, amplifying minor differences over cycles and explaining the essence of varying data distribution.

(3) A switching SOH estimation framework composed of LSTM and TemCap is put forward, enhancing the wide adaptability to unbalanced amounts of cycles in different stages.

(4) A transferable multi-stage SOH estimation model is proposed for batteries in the target domain, compensating for the deviation of cycling consistency of new batteries.

The remaining parts of this article are organized as follows: Section II describes LiBs data structure and mathematical formulation. The transferable multi-stage SOH estimation model is specified in Section III. The proposed method has been extensively compared to its counterparts through a benchmark dataset in Section IV. Finally, the conclusions and future works are given in the last section.

\section{Data Description and Problem Formulation}
This section illustrates the LiBs data structure and describes the transferable regression mathematically.
\subsection{Multi-stage Cycling Data Structure}
\subsubsection{Data Structure} Performance fading of LiBs includes multiple degradation behaviors over the whole degradation procedure, which is recognized as a multi-stage characteristic \cite{2020IECON_Multistage}. Typically, Jacqueline et al. \cite{2021Chemistry_Multistage} divides the life cycle into three main stages with the analysis of solid-electrolyte interphase: acceleration, stabilization, and saturation. The concept of stage describes the consecutive discharging cycles following the same degradation pattern. The degradation pattern maintains consistency within the same stage, but is dissimilar between different stages. This description is used to explain the transferable multi-stage idea, and specific stage division and online classification are beyond the scope of this work but can be found in the previous work \cite{2020IECON_Multistage}. Given there are $C$ stages in LiBs, the whole cycling data can be sequentially assigned into each stage accordingly. A three-dimensional data matrix $\bar{\textbf X}_c(I_c \times J \times K)$ is arranged for the cycling data collected in the $c^{th}$ stage, where $I_c$ is the number of cycles in the $c^{th}$ stage, $J$ is the number of variables, and $K$ is the length of samplings in each cycle. Correspondingly, the actual capacities of the battery in the $c^{th}$ stage are measured as $\mathbf Q_c \in \mathbb{R}^{1\times I_C}$.

Fig. 1 illustrates the degraded capacity with varying speeds provided by the National Aeronautics and Space Administration (NASA) \cite{2007NASA}. Essentially, the variable capacity is the external presentation of the inherent discharging signals. According to a prior knowledge \cite{2020IECON_Multistage}, the whole life span is divided into three stages, ranging from the healthy stage and slow-degradation stage to the rapid-degradation stage.

\begin{figure}[!htb]
\centering
\includegraphics[scale=0.48]{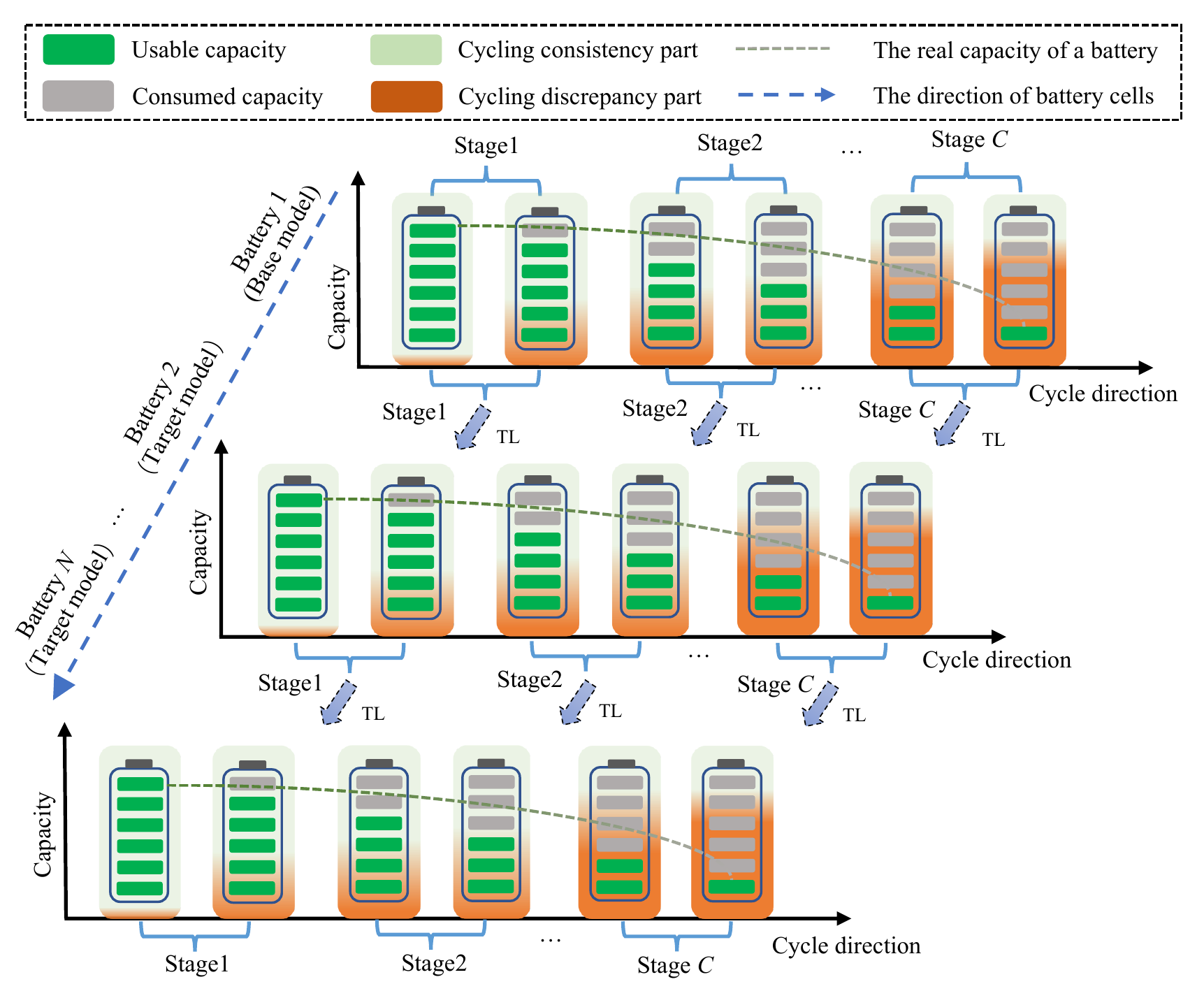}
\caption{The illustration of multi-stage transfer learning across batteries at the same stage for SOH estimation with cycling discrepancy analysis.}
\label{Fig2}
\end{figure}

\subsubsection{The Definition of SOH}
SOH is generally calculated by actual capacity over nominal capacity during cycling. Consequently, timely replacement of a LiB is highly recommended when the actual capacity is below a failure threshold, e.g., $70\%$ of its rated capacity \cite{2029TII_SOHCNNLSTM}. Typically, a fresh LiB is believed to be 100$\%$ healthy, and its health level will decline over cycles. Considering the stage information, the index $H_{c,i}$ is defined to indicate SOH in the $c^{th}$ stage as below,
\begin{equation}
H_{c,i} = {Q_{c,i}}/{Q} \in [0,1]
\label{con:Eq1}
\end{equation}
where $Q_{c,i}$ is current capacity of the $i^{th}$ cycle in the $c^{th}$ stage, and $Q$ is rated capacity for a battery.

\subsection{Problem Formulation for SOH Estimation with TL}
\subsubsection{Source Domain $\mathfrak{D}_S(\chi, P_S({\mathbf X}_{s}))$ in Stage $c$}
Source domain includes two critical parts: feature space $\chi$ and a marginal probability distribution $P_S({\mathbf X}_{s})$ about samples ${\mathbf X}_{s}$ from this domain. Given this, the purpose of source task is to find a regression $f_{TL}(\cdot)$ and a feature subspace $\chi$ at a certain stage from the source battery, which minimizes the error $E_s$ between the real capacities $\mathbf Q$ and predictions given below,
\begin{equation}
E_{s} = \sum \{\mathbf Q -  f_{TL} (\chi ({\mathbf X}_{s}))\}^2
\label{con:Eq2}
\end{equation}

\subsubsection{Target Domain $\mathfrak{D}_T(\mathbf x_T, f_{TL}(\cdot))$ in Stage $c$}
With $\chi$ and $f_{TL}(\cdot)$ from the source task, TL attempts to minimize the estimation error for cycle $\mathbf x_T$ from the target task. However, difficulty lies in that marginal probability distribution $P_T(\mathbf x_T)$ of target data is always different from $P_S({\mathbf X}_{s})$. Meanwhile, the amount of data in the target domain is insufficient to well-describe $P_T(\mathbf x_T)$. $f_{TL}(\cdot)$ derived in Eq. (\ref{con:Eq2}) expects to have a broad generalization ability, which could be used for the target battery at the same stage below,
\begin{equation}
Q_{T} = f_{TL} (\chi (\mathbf x_{T}))^2
\label{con:Eq3}
\end{equation}


Although cycling data like discharging voltage look similar, the available time-span is ever-decreasing over cycles because of the unavoidable influence of physicochemical reactions. In light of this, we use cycling consistency among cycles to stand for the similarity and the cycling discrepancy over cycles to indicate the distinctions. Fig. \ref{Fig2} visualizes the basic idea of cycling discrepancy analysis with TL in a three-dimension view, including capacity, stage, and battery.

It is worth pointing out that an assumption lies in the division of the number of stages for both source and target batteries. This assumption is made from the analysis of battery degradation mechanism \cite{2021MultiPhase}, as many batteries typically follow three states: health status, slight degradation, and rapid degradation. To delivery a coherent framework applied for most batteries, three stages is sufficient in this case.

\begin{figure*}[!htb]
\centering
\includegraphics[scale=0.3]{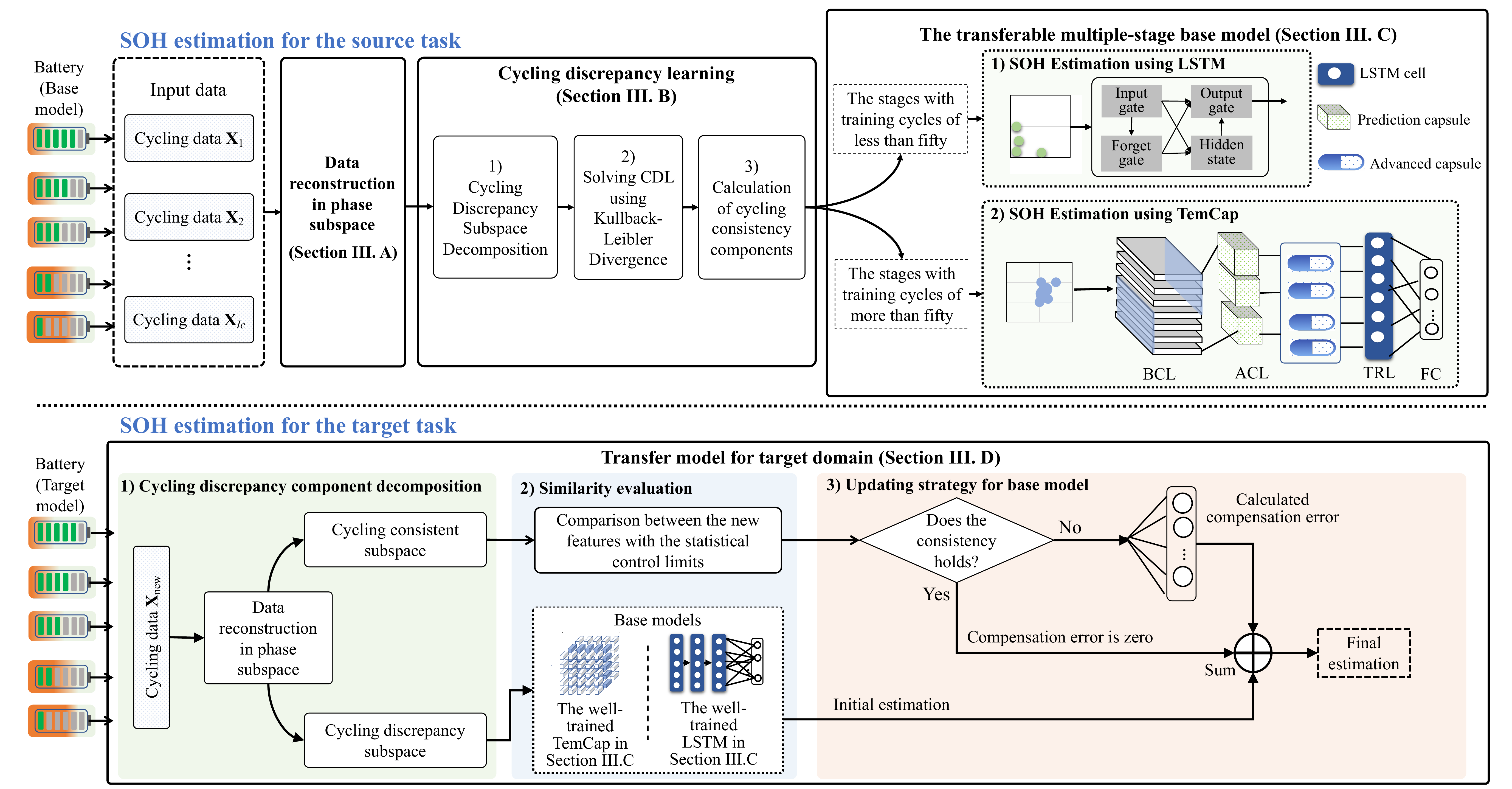} \\
\caption{Overview of the proposed transferable multi-stage model for LiB SOH estimation.}
\label{Fig3}
\end{figure*}

\section{A Transferable Multi-stage \\ SOH Estimation Model}
The proposed method is explained in this section, consisting of four sequential steps, as shown in Fig. \ref{Fig3}. First, data reconstruction explores in-depth dynamics from raw cycling data. Next, the cycling discrepancy learning finds the domain invariant representation to assist TL. Third, the classical LSTM and the proposed TemCap networks are comprehensively utilized for stages with different amounts of cycles. Last, a TL strategy for the target domain is proposed for online application purposes.

\subsection{Data Reconstruction in Phase Subspace}
\subsubsection{The Basic Idea of Phase Subspace}
Phase subspace theory originated from chaotic system analysis, and it is defined as a subspace where all possible states of the system are covered. The most important role of phase subspace allows for analyzing unseen system dynamics through a single observable time series. In this way, a specific state of the system can be represented by a point in the phase space and the time evolution of the system creates a trajectory in the phase space \cite{2020PhaseSpace}. As for the battery charging and discharging procedure, as shown in Fig. 4, phase subspace aims at describing the mechanics of lithium-ions through observable current, voltage, and temperature, linking the microstates to macrostates. With the flow of lithium-ions in microstates, the current and voltage are generated in macrostates. Although the clear description of the electrochemical reaction is difficult and even impossible, we apply phase space to macroscopic states of the degradation procedure by utilizing discharge current, discharge voltage, and environment temperature. A point in this phase space corresponds to a discharging cycle, with different available capacities as the initial condition.

\subsubsection{Degradation Data Reconstruction}
This part reconstructs raw cycling data grouped into each stage in a high-dimension space. In practice, discharge current, discharge voltage, and environment temperature are easily measured and widely used to infer battery capacity from a macroscopic perspective. LiBs involve physicochemical reactions that have complex microscopic properties, and it is impossible to reveal full dynamics only using current, voltage, and temperature. A solution is to describe the deep dynamics of a low-dimension system by exploiting a high-dimension embedding. This embedding is expected to uncover unknown and valuable information via the partially observable sensors. Phase subspace reconstruction (PSR) meets this purpose well, which is designed to reveal the underlying dynamics of a system with partial low-dimension measurements \cite{XU2009PSR}. In addition, data reconstruction can graphically detect hidden patterns and dissimilarities across the cycling data.

With dimensions of reconstructed space $r$ and time delay $\tau$, $\mathbf X_i$ is extended to a high-dimension matrix below,
\begin{equation}
\begin{aligned}
\mathbf{X}_i =\left[\mathbf{V}_{i}, \mathbf{I}_{i}, \mathbf{T}_{i}\right] \quad i \in [1,C]\\
\end{aligned}
\label{con:Eq4}
\end{equation}
where,
\begin{equation}
\begin{aligned}
\scriptsize
\mathbf{V}_i=\left\{\begin{array}{llll}
V_{i}(1) & V_{i}(1+\tau)  & \cdots & V_{i}(1+(r-1) \tau) \\
V_{i}(2) & V_{i}(2+\tau)  & \cdots & V_{i}(2+(r-1) \tau) \\
\cdots & \cdots & \cdots & \cdots \\
V_{i}(k) & V_{i}(k+\tau)  & \cdots & V_{i}(k+(r-1) \tau) \\
\cdots & \cdots & \cdots & \cdots \\
\end{array}\right\}
\end{aligned}
\end{equation}
and $\mathbf X_i$ is the $i^{th}$ cycle of $\bar{\textbf X}_c$; $\mathbf V_{i}$ is the reconstructed data for voltage; the reconstructed current $\mathbf I_{i}$ and temperature $\mathbf T_{i}$ can be obtained in a similar way.

$r$ influences the dimension of reconstructed space, which can be computed using the false neighbor method \cite{XU2009PSR}. A proper $\tau$ is determined when mutual information between $\mathbf x$ and its lagged vector is minimal. Performing Eq. (\ref{con:Eq4}) to each degradation cycle in the same stage, we will obtain the number of $I_C$ reconstructed cycling data using PSR. For clarity, the reconstructed data are still denoted as ${\mathbf X}_1$, ${\mathbf X}_2$, to ${\mathbf X}_C$ corresponding to Cycles 1 to $I_C$, and the dimension of the reconstructed space is still denoted as $J$.

\subsection{Cycling Discrepancy Learning}
This part separates the cycling discrepancy components from the reconstructed data in each stage using cycling discrepancy learning (CDL).
\subsubsection{Cycling Discrepancy Subspace Decomposition}
According to the above analysis, $\mathbf X_i$ can be the superposition of cycling consistency $\mathbf S_{i,s}$ and cycling discrepancy $\mathbf S_{i,d}$ when a linear mapping relationship is imposed. Through separating $\mathbf S_{i,d}$ from $\mathbf S_{i,s}$, invertible  matrices $\mathbf P_{s}$ and $\mathbf P_{d}$ are corresponding subspaces, to derive the cycling discrepancy and consistency components. Then we have,
\begin{equation}
[\begin{array}{l}
\mathbf S_{i,s} \\
\mathbf S_{i,d}
\end{array}] = \mathbf P {\mathbf X_i}^T =
[\begin{array}{l}
\mathbf P_s \\
\mathbf P_d
\end{array}] {\mathbf X_i}^T
\label{con:Eq6}
\end{equation}

Essentially, this problem aims to find the cycling discrepancy features $\mathbf S_{i,d}$. However, $\mathbf S_{i,d}$ is always non-stationary along cycle-wise direction. As such, $\mathbf P_{d}$ is subject to the nonunique solutions when Eq. (\ref{con:Eq6}) is directly solved. An alternative solution is to locate $\mathbf S_{i,s}$, which almost keeps consistent among cycling data. Consequently, the cycling discrepancy components can be derived by projecting the cycling data onto the orthogonal subspace. The cycling consistency components in Cycle $i$ are decomposed according to Eq. (\ref{con:Eq6}) below,
\begin{equation}
\mathbf{S}_{i,s}=\mathbf{P}_{s} {\mathbf X_i}^T
\label{con:Eq7}
\end{equation}

\subsubsection{Solving CDL using Kullback-Leibler Divergence (KLD)}
Projecting $\mathbf X_1$ through $\mathbf X_{I_C}$ onto the space $\mathbf {P}$, a series of features will be computed as $\mathbf S_1, \mathbf S_2, \cdots, \mathbf S_{I_C}$. Since it is a considerable burden to evaluate the consistency between these features one by one, an efficient way is to employ Gaussian distribution with zero mean and unit variance as an auxiliary indicator, a typical stationary signal. Then KLD is employed to measure the distribution divergence between $\mathbf S_{i,s}$ and Gaussian distribution. As such, the optimal projection matrix $\mathbf P_{s}$ corresponds to the minimal KLD.

With Z-score normalization, the mean and variance of $\mathbf X_i$ are calculated as $\mathbf u_i$ and $\mathbf \Sigma_i$, respectively. Introducing a whitening matrix $\mathbf W_c$ as a part of $\mathbf P$, the cycling consistency components are obtained below,
\begin{equation}
\mathbf{S}_{i,s}= \mathbf I_s \mathbf P {\mathbf{X}_i}^T= \mathbf I_s \mathbf{\Pi W}_c {\mathbf{X}_i}^T
\label{con:Eq8}
\end{equation}
where $\mathbf \Pi$ is the updated projection matrix to be solved, $\mathbf I_s$ is the identity matrix $\mathbf I(J \times J)$ truncated to the first $S$ rows, and $\mathbf W_c$ is calculated by $\sqrt{\mathbf{\tilde X}_c^T \mathbf {\tilde X}_c}$, where $\mathbf{\tilde X}_c$ is the average of all cycling data in the $c^{th}$ stage.

By performing linear transformation $\mathbf I_s \mathbf{\Pi W}_c {\mathbf{X}_i}^T$ on each cycling data, solving Eq. (\ref{con:Eq8}) is transformed to an optimization problem by minimizing the objective function as follows,
\begin{equation}
\begin{aligned}
L(\mathbf{\Pi}) &=\sum_{i=1}^{I_C} D_{KL}\left[N\left(\tilde{\mathbf{u}}_{i,s}, \tilde{\mathbf{\Sigma}}_{i, s}\right) \| N(0, \mathbf{I})\right] \\
&=\sum_{i=1}^{I_C} \sum_{m} P_{i,s}(m) \log \left(\frac{P_{i,s}(m)}{Q(m)}\right)
\end{aligned}
\label{con:Eq9}
\end{equation}
where $\tilde{\mathbf{u}}_{i,s}=\mathbf I_s \mathbf {\Pi W}_c \mathbf u_i$ and $\tilde{\mathbf{\Sigma}}_{i, s}= \mathbf I_s \mathbf {\Pi W}_c \mathbf \Sigma_i (\mathbf I_s \mathbf {\Pi W}_c)^T$ are the mean and variance after linear transformation, respectively; $D_{KL}(\cdot)$ is the KLD function; $N(0, \mathbf I)$ is Gaussian distribution with zero mean and unit variance; $P_{i,s}(m)$ is the probability density of $N(\tilde{\mathbf{u}}_{i,s} ,\tilde{\mathbf \Sigma}_{i,s})$ for the $m^{th}$ sample; $Q(m)$ is the probability density of $N(0, \mathbf I)$ for the $m^{th}$ sample.

\begin{figure}[!htb]
\centering
\includegraphics[scale=0.35]{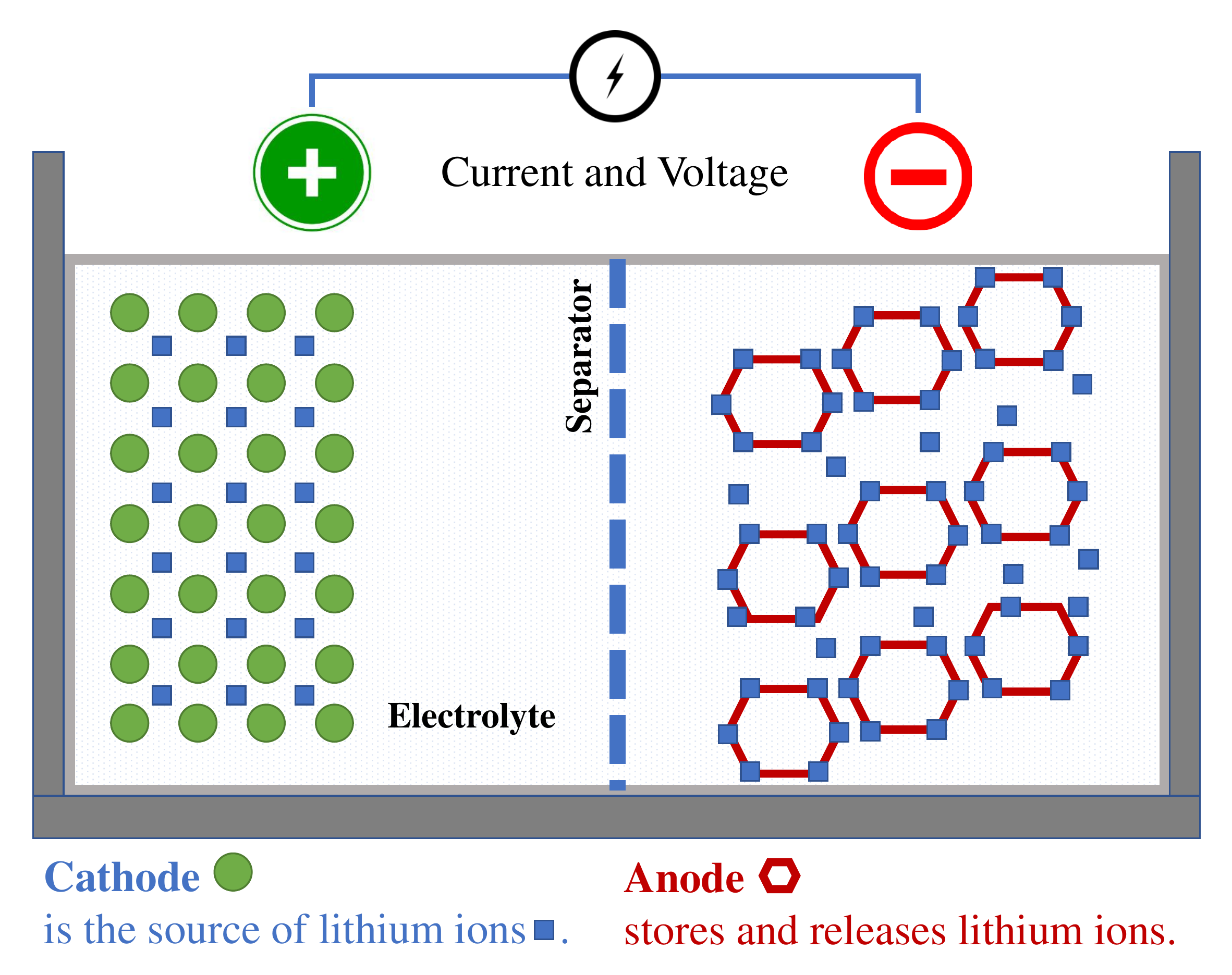} \\
\caption{The microstate movements of LiB with lithium ions and the macrostates with current and voltage.}
\label{Fig4}
\end{figure}

\subsubsection{Calculation of Cycling Consistency and Discrepancy Components}
The matrix $\mathbf \Pi$ is solved by conjugate gradient descend [13]. Cycling consistency components with dimensions $S \times K$ are derived below,
\begin{equation}
\mathbf{S}_{i,s}=\mathbf{I}_{s} \mathbf{\Pi W}_c {\mathbf{X}_i}^T= \mathbf{\Pi}_{s} \mathbf{W}_c {\mathbf{X}_i}^T
\label{con:Eq10}
\end{equation}

Consequently, the cycling discrepancy components  with dimensions $(J-S) \times K$ are spanned by the remaining subspace $(\mathbf \Pi - \mathbf \Pi_s)$. Then the components $\mathbf S_{i,d}$ in Cycle $i$ can be estimated as follows,
\begin{equation}
\mathbf{S}_{i, d}=(\mathbf \Pi - \mathbf{\Pi}_{s}) \mathbf{W}_c {\mathbf{X}_i}^T
\end{equation}

\subsection{The Transferable Multiple-stage Source Model}
The multi-stage source model has been developed with the proposed TemCap and a classical LSTM. Combing these two models is due to the unbalanced distribution of cycles over stages. Although the proposed TemCap shows better feature representation compared to LSTM, it requires more data for model training. Usually, for a typical three-stage degradation procedure, the number of cycles in the first and third stages is less than that of the second degradation stage. As a rough guide, we suggest the LSTM-based estimation model for stages with less than 50 training cycles. As such, TemCap can be applied for stages with more than 50 training cycles.

\subsubsection{SOH Estimation using LSTM}
LSTM has been widely applied in various industrial fields for sequential learning, such as risk prediction \cite{TIIEditor1} and performance prediction \cite{2021Editor4}. Its temporal learning ability is ensured through elaborately designed LSTM cells, which possess memory capability and overcome the vanishing gradient problem.


Taking $\mathbf S_{i,d}$ as the input, LSTM layers integrate and compress cycling data into the final hidden state for regression. An essential of the LSTM cell is the introduction of the memory and forget mechanism. By storing historical information into the hidden state $\mathbf h_{k-1}$, the candidate memory mechanism removes useless information for regression and retains the essential part. Next, $\mathbf h_k$ at time $k$ is iteratively updated with the information from input gate, forget gate, and output gate at the next time, and then essential information contained in $\mathbf S_{i, d}$ will be compressed into the final vector. Let $\mathbf H_{i, d}$ be the collection of output vectors of the LSTM layer. This flattened information passes through a fully connected layer below,
\begin{equation}
\mathbf {\hat Q} = F_f\left(\mathbf W_f \mathbf H_{i,d}+ \mathbf B_f \right)
\label{con:Eq14}
\end{equation}
where $\mathbf {\hat Q}$ represents the predictions; $\mathbf W_f$ and $\mathbf B_f$ represent the transformation weights and bias of fully connected layer, respectively; and ReLu function is usually selected for $F_f(\cdot)$.



\subsubsection{SOH Estimation using TemCap}
A TemCap model is proposed to enable deep feature representation with a three-layer hierarchical structure for stages with sufficient training data. The first two layers focus on feature representation within each cycle. Specifically, to better present the overall data in a cycle, the part-whole relationship is introduced by breaking the entire cycle into a series of local representations. In the third layer, temporal correlations among cycling data are linked with each other. The information density is from low to high, corresponding from the first layer until the third layer.

The designed hierarchical structure fully utilizes information of discharging data within a cycle and over cycles. To achieve this purpose, we use basic capsules to describe the local information and advanced capsules to indicate the global information. A capsule is  defined originally as the fundamental element for feature representation, which has been used to describe properties such as pose, texture, or deformation of an entity or a part of an entity \cite{KWABENAPATRICK2019}. Mathematically, it is a multidimensional vector achieved through a group of neurons, each of which stands for one dimension. The details of each layer and its formulations are described below:

\textit{$\bullet$ Basic Capsule Layer (BCL):} Basic capsules are designed to capture the local features of cycling data, which could be denoted as a vector $\mathbf{u}_m$ for the $m^{th}$ capsule. From the viewpoint of feature representation, capsules are the basic units in the feature space, and their combinations perform high-level features. Projecting $\mathbf S_{i,d}$ calculated in Eq. (10) into a space ${\mathbf \Xi}$, a series of feature maps are calculated below,
\begin{equation}
\mathbf U = f({\mathbf \Xi} {\ast} \mathbf S_{i,d} + \mathbf b)
\label{con:Eq15}
\end{equation}
where $\ast$ is convolutional operation, $f(\cdot)$ is the ReLu activation function, $\mathbf b$ is bias, and $\mathbf U$ is collection of basic capsules.

$\mathbf{\Xi}$ determines the quality of generated local features in an iterative learning way. Initially, $\mathbf{\Xi}$ can be generated through convolutional neural layers with randomly generated convolution kernels, but a pooling operation is emitted to avoid information loss. A basic capsule is constructed by grouping values at the same position in feature maps with a length $M$. As such, the continuous $M$ kernels correspond to a certain space $\mathbf \Xi$. A series of basic capsules with dimensions $1 \times M$ will be constructed. This layer transforms the scalar-based outputs into vector-based outputs with the capsule concept.

\begin{figure*}[!ht]
	\centering
	\subfigure[]
	{
	\begin{minipage}[t]{1\linewidth}
	\centering
	\includegraphics[width=18cm]{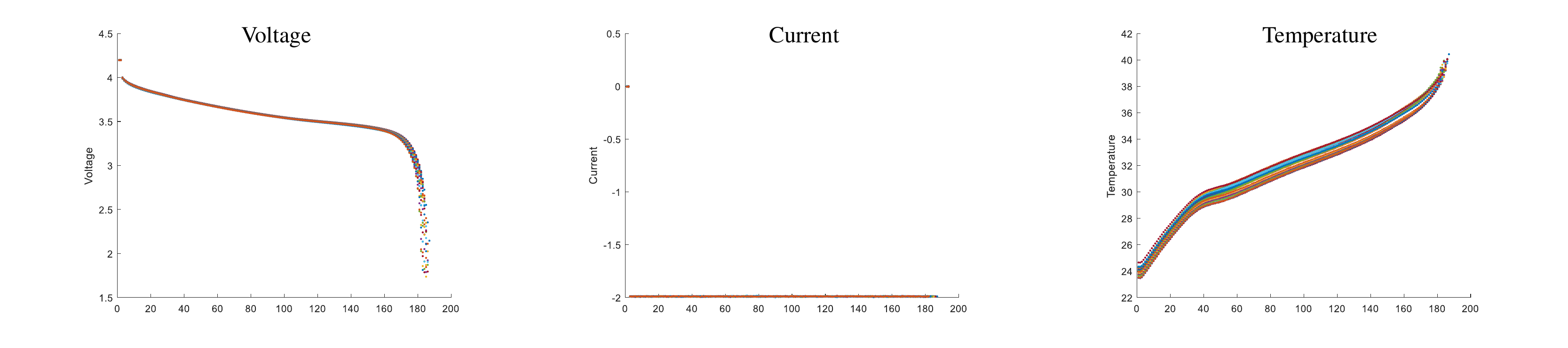}
	\end{minipage}
	}
	
	\subfigure[]
	{
	\begin{minipage}[t]{1\linewidth}
	\centering
	\includegraphics[width=18cm]{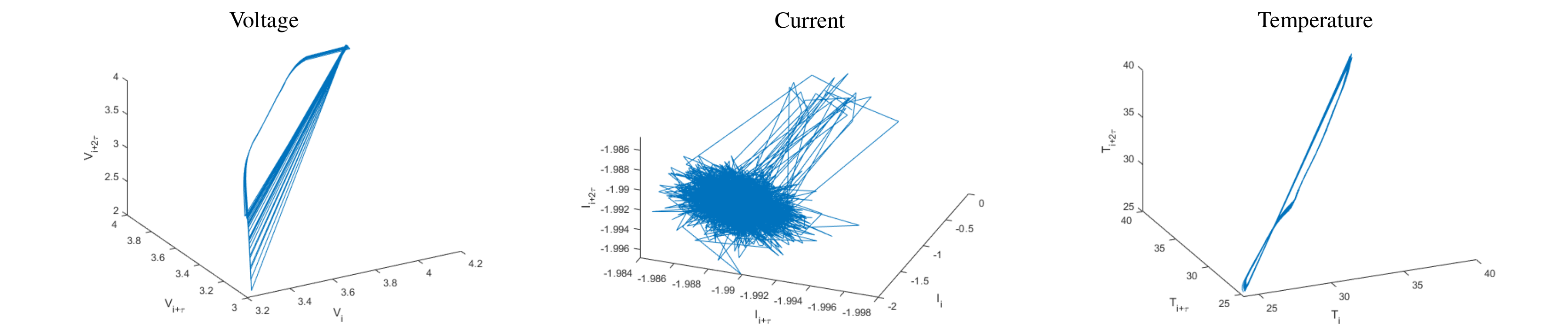}
	\end{minipage}
	}
	\caption{Visualization comparison of (a) raw cycling data and (b) corresponding reconstructed data in Stage 1 of Battery B7.}
\label{Fig5}
\end{figure*}

\textit{$\bullet$ Advanced Capsule Layer (ACL):} Advanced capsules are high-dimensional features conveying stronger representations compared to basic capsules. For instance, for classification problems, advanced capsules allow distinguishing a dog from a cat by simultaneously viewing the orientation of eyes, nose, ears, mouse, etc. The input of an advanced capsule $j$ is calculated by summarizing all basic capsules below,
\begin{equation}
\mathbf v_{j}=\sum_{m} c_{m, j} \mathbf W_{m, j} \mathbf u_{m}
\label{con:Eq16}
\end{equation}
where $\mathbf W_{m, j}$ is the projection matrix to extend low-dimension space to high-dimension space, and $c_{m, j}$ is the coefficient.

As the similar expression of basic capsules, we use a neuron to stand for one dimension of an advanced capsule, and a complete advanced capsule consists of several neurons corresponding to the given dimension. Sequential neurons in fully connected layers are grouped to construct advanced capsules. Essentially, we need to find a high-dimensional space spanned by the advanced capsules.  Besides, as the integration of basic capsules distributed at a specific time in a discharging cycle, the advanced capsules contain temporal information.

\textit{$\bullet$ Temporal Representation Layer (TRL):} This layer focuses on acquiring temporal information over cycles, linking advanced capsules within each cycle. It allows to receive capacity fading trend over cycles and estimate the possible values with the advanced capsules. As such, we stack all advanced capsules of the $i^{th}$ cycle into one dimension vector $\mathbf{V}_i=[\mathbf{v}_1, \mathbf{v}_2, \dots, \mathbf{v}_D]$ using the flatten operation. Then the collection of advanced capsules in each cycle is sequentially fed into LSTM cells-based regression model. For clarity, the details of LSTM are not repeated here.

\textit{$\bullet$ Weight Learning for TemCap:} The network parameters include projection subspace $\mathbf{\Xi}$ in the basic capsule layer, projection subspace in the advanced capsule layer, LSTM cells, and the connected links between each layer. The proposed TemCap is trained iteratively, which integrates both dynamic routing mechanism and gradient descent to update the network's weights and bias.

We employ the classical dynamic routing \cite{2017CapsNet} to update the links between basic capsules and advanced capsules below,
\begin{equation}
\mathbf s_{j}=\frac{\|\mathbf v_{j}\|^{2}}{1+\| \mathbf v_{j}\|^{2}} \frac{\mathbf v_{j}}{\|\mathbf v_{j}\|}
\label{con:Eq17}
\end{equation}
where $\mathbf s_j$ is the output of $j^{th}$ advanced capsule after updating, and $\mathbf v_j$ has the same meaning as that in Eq. (\ref{con:Eq16}).

From Eq. (\ref{con:Eq17}), $\mathbf v_j$ with a small length will shrink to slightly above zero. In contrast, $\mathbf v_j$ with a large length approaches approximately one. As such, length of the output vector of a capsule indicates the probability of a local feature. Dynamic routing is mainly used for optimizing parameters of basic capsules and advanced capsules. The remaining parameters, including projection subspace to derive basic capsules and advanced capsules, and LSTM layers, can be updated by Gradient descent.

\subsection{Transfer Model for Target Domain}
This section discusses the transfer strategy when the well-developed source models are adopted for other LiBs.

\begin{figure*}[!ht]
\centering
\subfigure[]
{
\begin{minipage}[t]{0.3\linewidth}
\centering
\includegraphics[width=6cm]{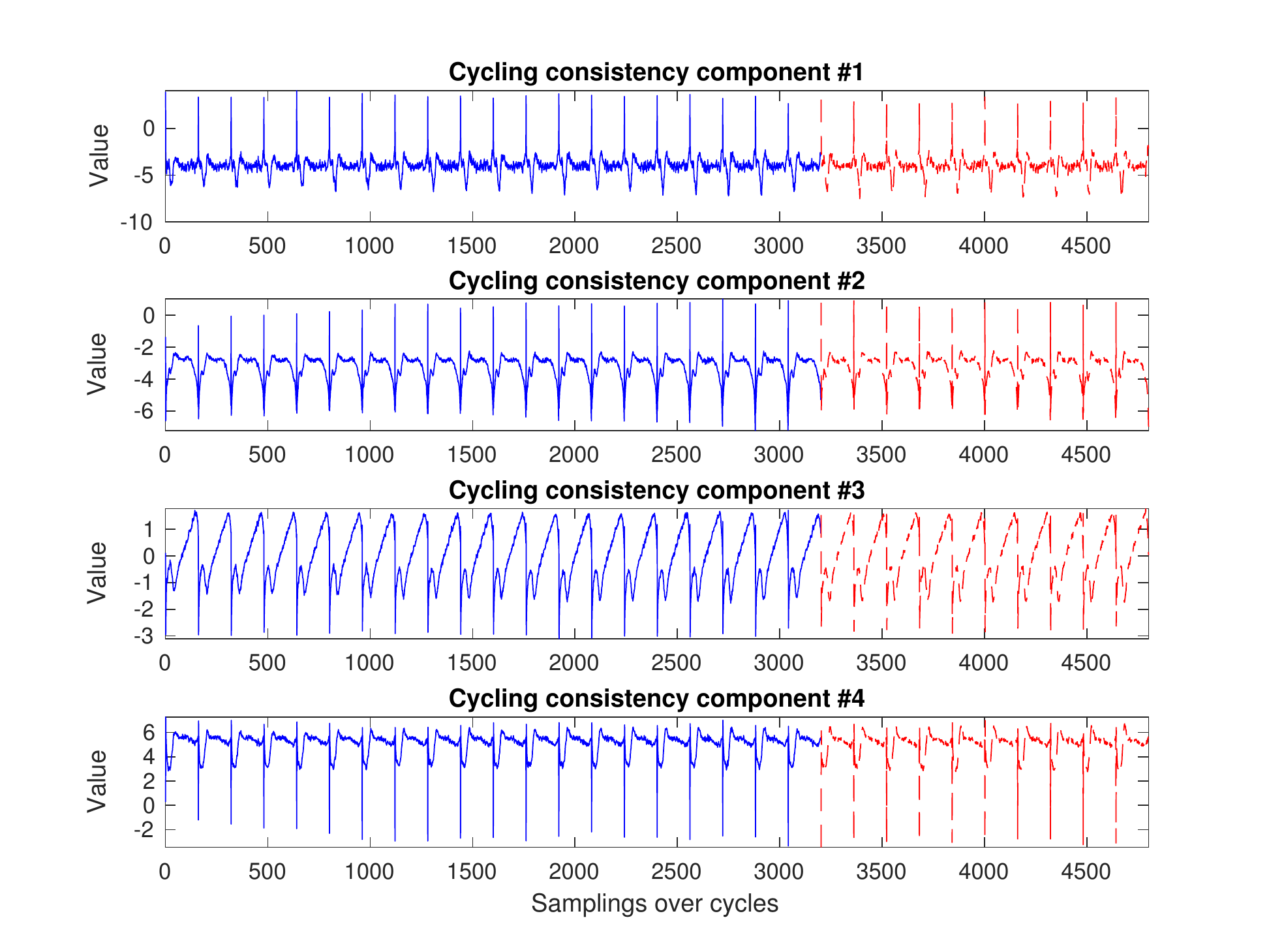}
\end{minipage}
}
\subfigure[]
{
\begin{minipage}[t]{0.3\linewidth}
\centering
\includegraphics[width=6cm]{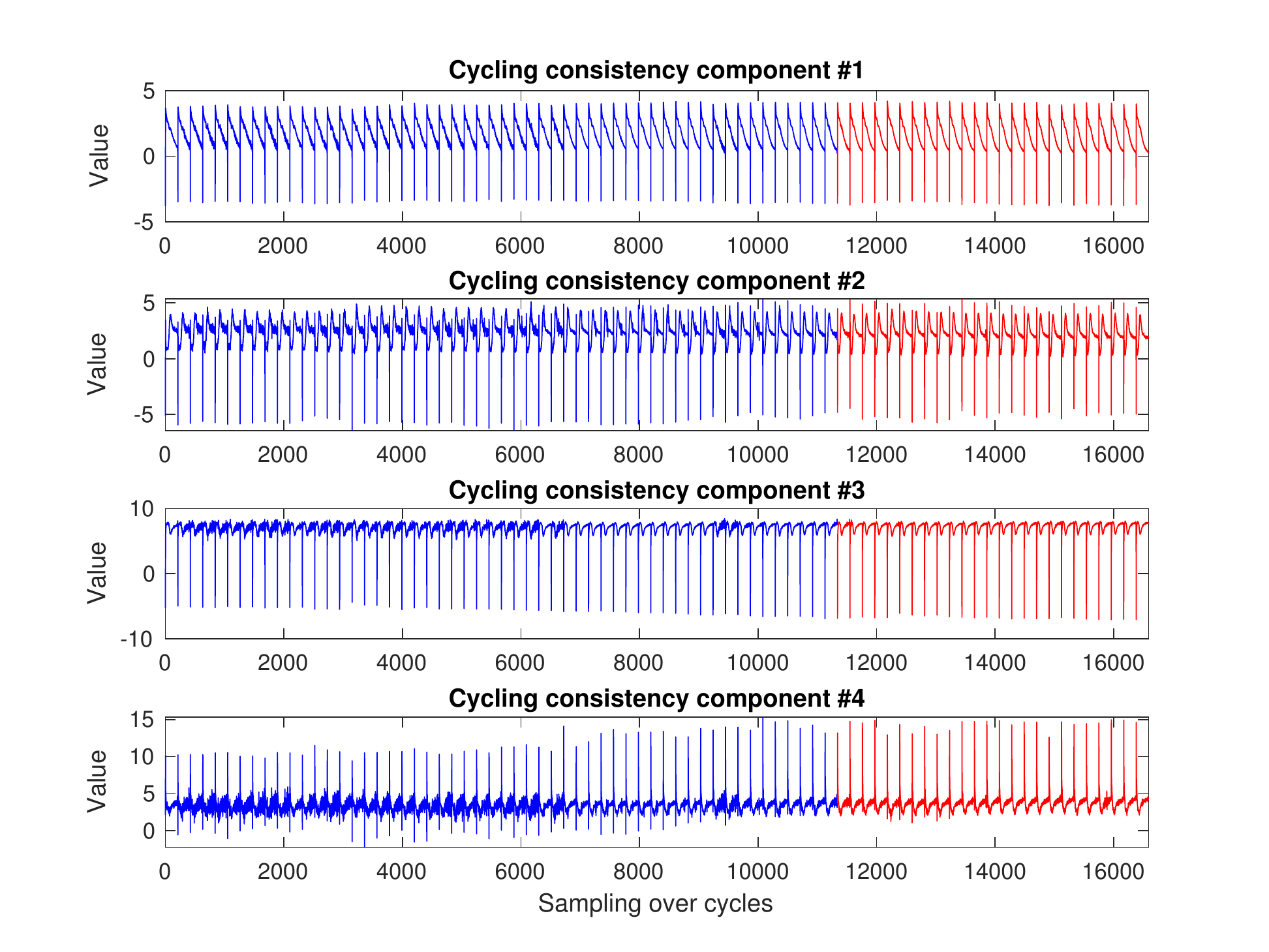}
\end{minipage}
}
\subfigure[]
{
\begin{minipage}[t]{0.3\linewidth}
\centering
\includegraphics[width=6cm]{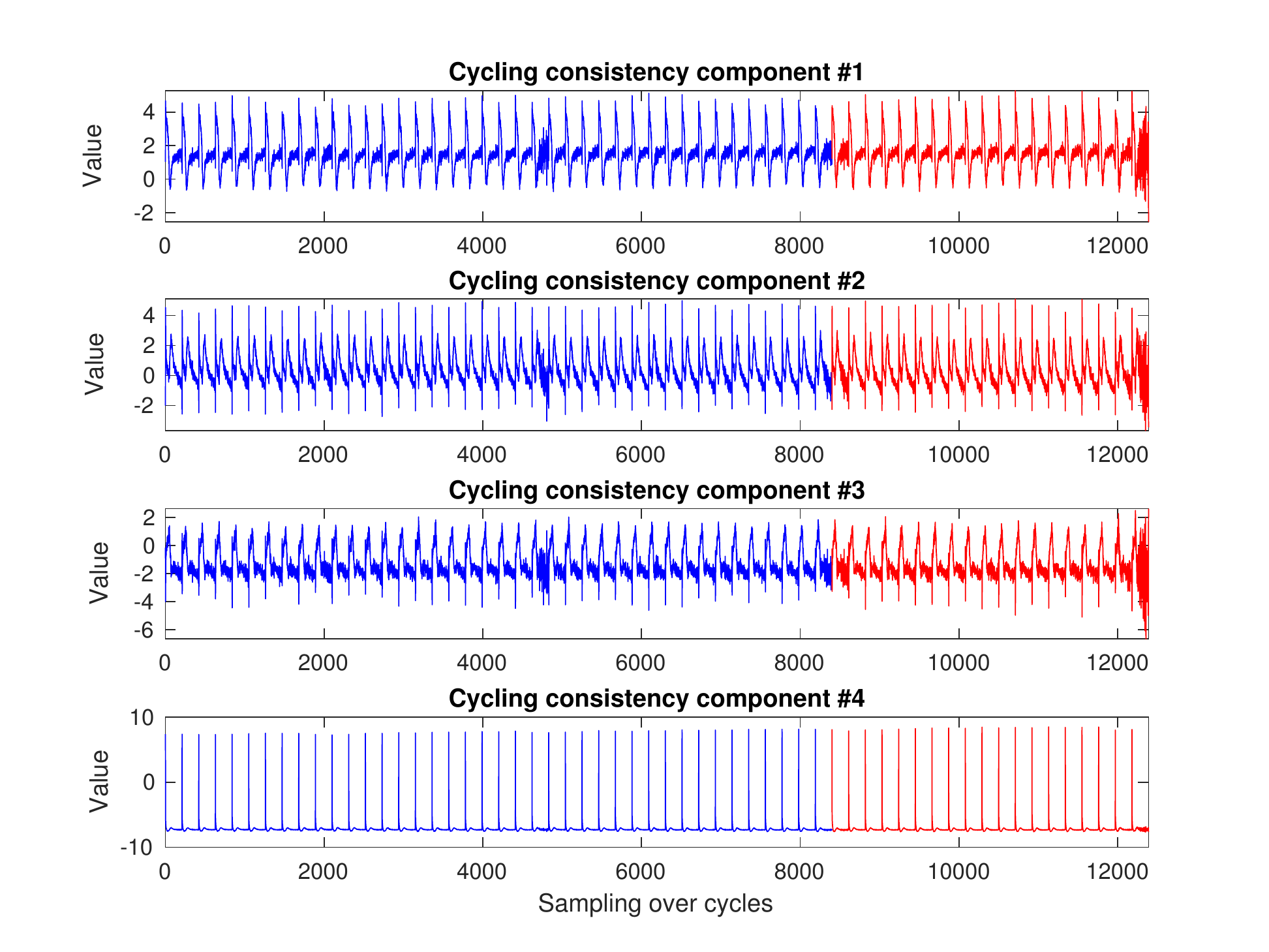}
\end{minipage}
}
\caption{Visualization of decomposed cycling consistent features with known stage division for Battery B7 in (a) Stage 1, (b) Stage 2, and (3) Stage 3.}
\label{Fig6}
\end{figure*}

\begin{figure*}[!ht]
\centering
\subfigure[]
{
\begin{minipage}[t]{0.3\linewidth}
\centering
\includegraphics[width=6cm]{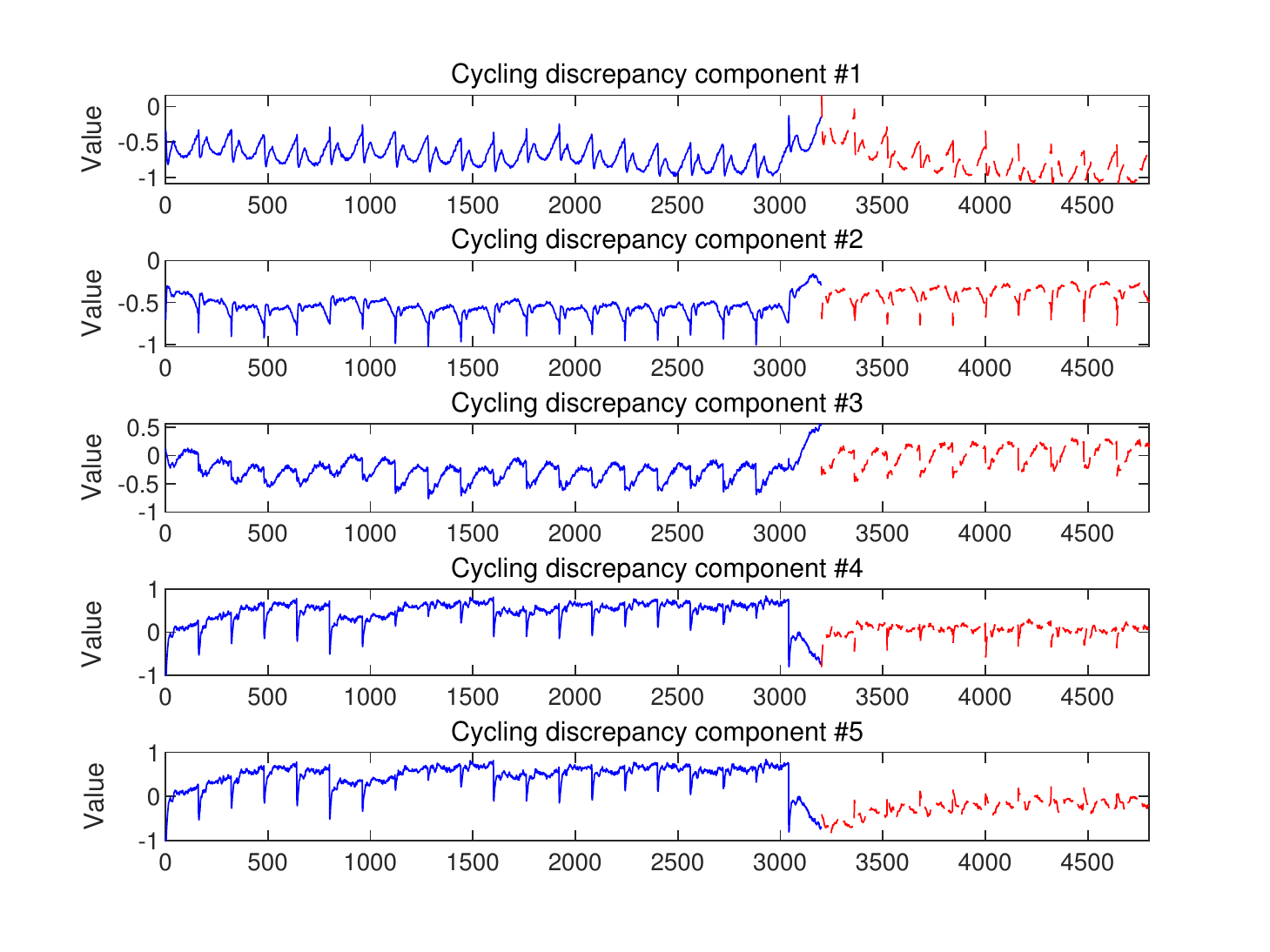}
\end{minipage}
}
\subfigure[]
{
\begin{minipage}[t]{0.3\linewidth}
\centering
\includegraphics[width=6cm]{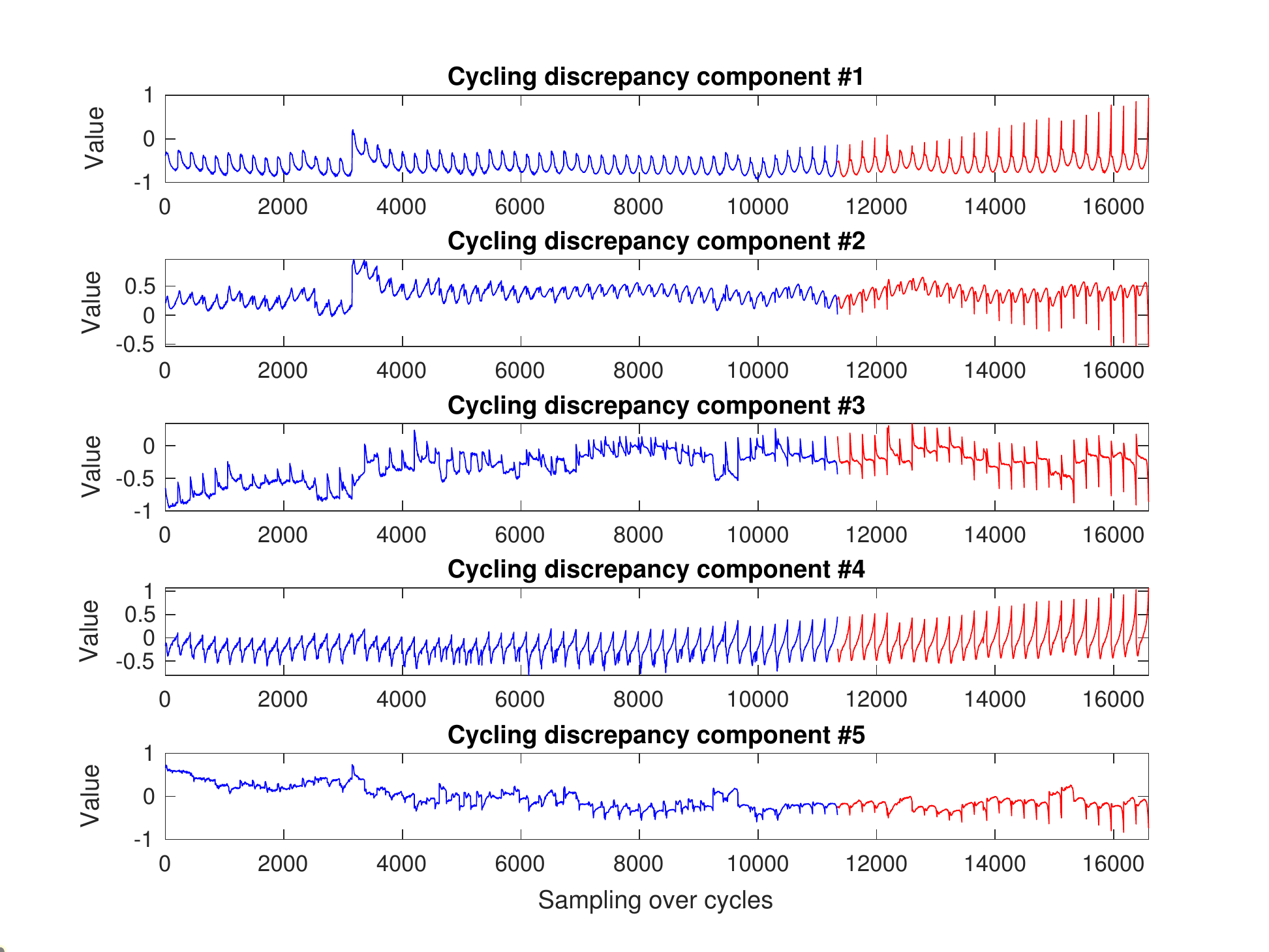}
\end{minipage}
}
\subfigure[]
{
\begin{minipage}[t]{0.3\linewidth}
\centering
\includegraphics[width=6cm]{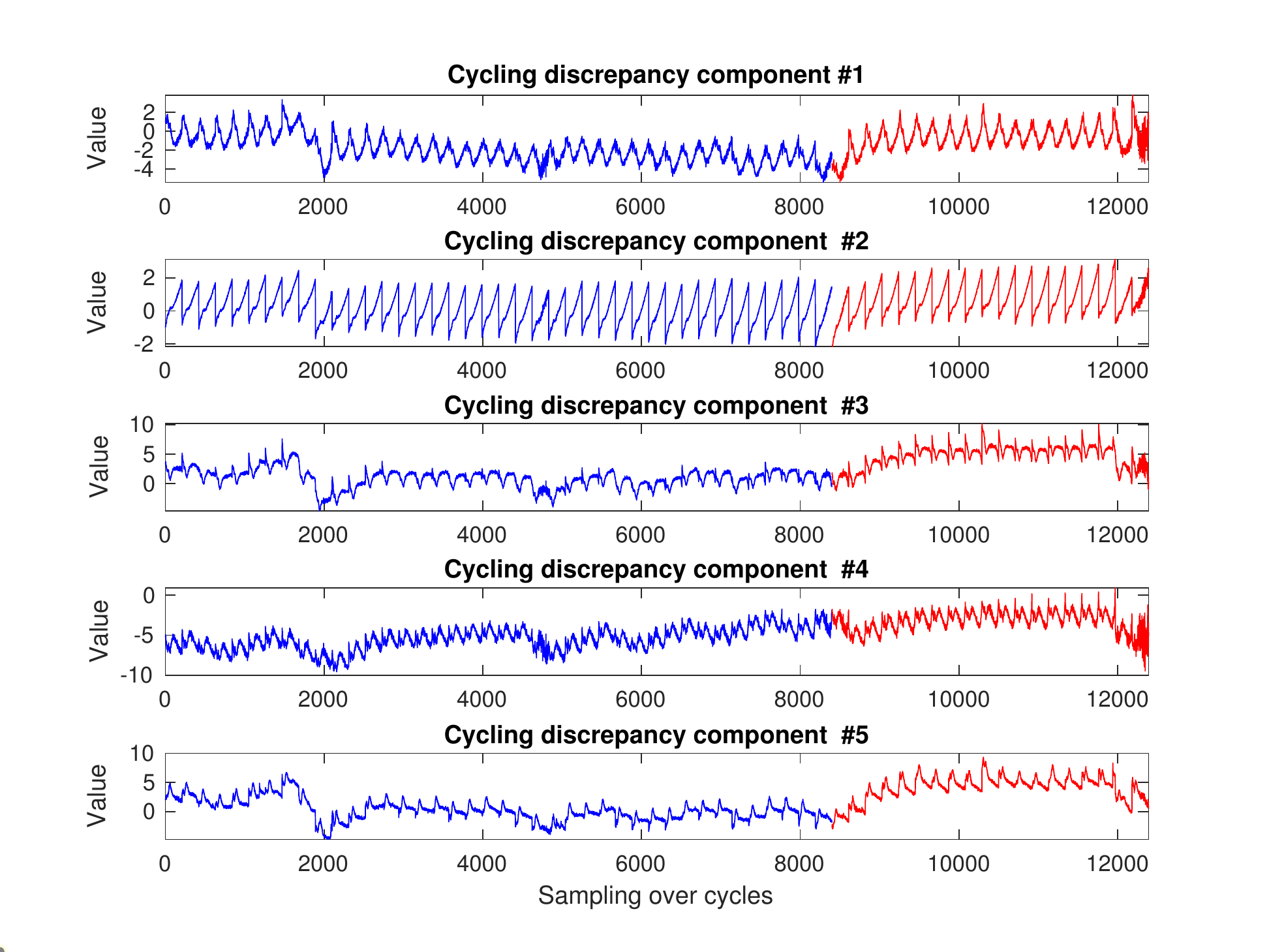}
\end{minipage}
}
\caption{Visualization of decomposed cycling discrepancy features with known stage division for Battery B7 in (a) Stage 1, (b) Stage 2, and (3) Stage 3.}
\label{Fig7}
\end{figure*}

\subsubsection{Cycling Discrepancy Component Decomposition}
When an online reconstructed cycling matrix $\mathbf X_{new}$ is available, its belonging to a specific stage is identified by evaluating the stage cycling consistency \cite{2020IECON_Multistage}. Afterwards, cycling consistency components $\mathbf{S}_{new,s}$ and cycling discrepancy components $\mathbf{S}_{new,d}$ of $\mathbf X_{new}$ are decomposed below,
\begin{equation}
\begin{array}{l}
\mathbf{S}_{new,s} = \mathbf{\Pi}_{s} \mathbf{W}_c \mathbf{X}_{new}^T \vspace{1.5ex} \\
\mathbf{S}_{new,d} = (\mathbf \Pi - \mathbf{\Pi}_{s}) \mathbf{W}_c \mathbf{X}_{new}^T
\end{array}
\end{equation}



\subsubsection{Similarity Evaluation}
Ideally, the well-trained source model expects to gain similar accuracy when it aims at the target task. Considering that the derived cycling discrepancy reflects the differences among LiBs caused by natural reaction and the varying operating conditions, the defined similarity analysis contributes to judge whether the developed source model in a specific stage can be directly applied considering manufacturing differences in the consistent phase. Naturally, this kind of manufacturing difference performs constant behaviors during the capacity fading procedure and influences the consistent features over cycles. The stable behaviors may affect either all stages or partial stages, which inspires us to conduct similarity checking from stage to stage. We establish the control limit for cycling consistency components $\mathbf S_{i,s}$ from the training dataset, which follows the $F$ distribution below,
\begin{equation}
CL_s=\mathbf{\bar S}_{s} \mathbf \Lambda_{s}^{-1} \mathbf{\bar S}_{s}^T \sim \frac{S \left(N_c^{2}-1\right)}{N_c(N_c-1)} F_{S, N_c-S, \alpha_s}
\end{equation}
where $\mathbf \Lambda_{s}$ is the covariance matrix of $\mathbf{\bar S}_{s}$, which is the average of all components $\mathbf S_{i, s}$ in a certain stage, $\alpha_s$ is the significant level (here is 0.05) to derive the 95$\%$ confidence limit, and $N_c$ is the total number of samples in Stage $c$.

If $\mathbf S_{new,s}$ is covered by the control limit $CL_s$, the developed source models could be directly adopted by feeding $\mathbf S_{new,d}$. Otherwise, we will update the compensation information fromn the target data into the source model.

\subsubsection{Updating Strategy for Source Model} The updating strategy allows to rapidly transfer the well-developed source model from a battery cell to others, handling the estimation error caused by the steady discrepancy between the source and target domains. A handful of discharging cycles at the beginning of the stage are denoted as $\mathbf X_{new, 1}$ to $\mathbf X_{new, T}$. The correspondingly actual capacities are remarked as ${\mathbf Q}_T$, and then their predictions $\hat {\mathbf Q}_T$ of the original source model can be calculated. The vector of estimation error is computed as $\mathbf {E}_T = \hat {\mathbf Q}_T - \mathbf {Q}_T$.

A shallow neural network with a fully connected layer is adopted to regress $\mathbf E_T$ on the decomposed consistent components $\mathbf  S_{new,d}$. In a parallel structure, final prediction for the target task is the sum of the prediction from the source model and the error compensation from the shallow neural network.

\textit{Remark:} The training time complexity of the proposed method is discussed, mainly focusing on two kinds of neural networks. First, for the lightweight LSTM network, the overall time complexity per time step is $o(W)$ \cite{2017TimeComp}, where $W$ is the number of all weights in the network. Second, for TemCap, the time complexity of the basic capsule layer is $o(n_{b} \times s_b^2 \times m_b^2)$ \cite{2018TimeComp}, where $n_b$ is the number of filters, $s_b$ is the spatial size of the filter, and $m_b$ is the spatial size of the output feature map. The time complexity of the advanced capsule layer is calculated as $o(M \times n \times m)$, where $M$ is the dimension size of a basic capsule, $n$ is the number of the basic capsule, and $m$ is the number of advanced capsules. With an LSTM layer, the total time complexity during training of TemCap is $o(n_{b} \times s_b^2 \times m_b^2 + M \times n \times m + W)$.

\section{Experimental Results and Discussions}
The efficacy of the proposed method is verified in this section, where a series of comparisons have been conducted to exhibit the advantages of data reconstruction, cycling discrepancy analysis, and the proposed SOH estimation model.

A widely used benchmark dataset from the NASA repository is employed for experiments, including the short-term cycling stage and the long-term cycling stage \cite{2007NASA}. Three battery cells, i.e., B5, B6, and B7, are selected for comprehensive analysis and comparisons. They are 18650 LiB with a rated capacity of around 2Ah. Operating with a 2A constant current mode at room temperature of 24$^{\circ}$C, a discharging cycle stops when the battery voltages drop to 2.7V, 2.5V, and 2.2V, respectively. During each discharging cycle, discharging voltage, discharging current, and environment temperature will be measured for analysis. We borrow the general stage division results from \cite{2020IECON_Multistage}. As such, all discharging cycles of each battery have been divided into three stages for easy online implementation, as listed in Table I.

\begin{table}[!htb]
\caption{Multi-stage division information for employed LiBs.}
\renewcommand{\arraystretch}{1.5}
\centering
\scriptsize
\begin{center}
\begin{threeparttable}
\begin{tabular}{c| c| c| c| c}
\hline
\hline
{\textbf {Name}} & {\textbf {Total cycles}} & {\textbf {Stage 1}} & {\textbf {Stage 2}} & {\textbf {Stage 3}} \\
\hline
{B5, B6, B7} & 167 & 1-30 & 31-106 & 107-167  \\
\hline
\hline
\end{tabular}
\end{threeparttable}
\end{center}
\end{table}

The index root mean square error ($RMSE$) is adopted to evaluate the accuracy of SOH estimation result, which is defined as follows,
\begin{equation}
RMSE_{SOH}(\%) = \frac{1}{Q} \sqrt{\frac{1}{I_C} \sum_{i=1}^{I_c} (\hat{Q}_{i, T} - Q_{i, T})^2} \times 100(\%)
\end{equation}
where $Q$ is rated battery capacity, $\hat{Q}_{i,T}$ denotes the estimated value of $i^{th}$ cycle of the target battery, ${Q}_{i,T}$ is the corresponding ground truth, and $I_C$ is the number of testing cycles.

Fair comparisons are ensured through the same hardware and software sources. Specifically, all experiments were conducted on an HP workstation equipped with a CPU Intel Xeon E5-2620 v4 CPU (16 cores and each of them is 2.10 GHz), 32GB RAM, and a graphics processing unit (NVIDIA GeForce GTX 1080 Ti). The LSTM model is programmed by MATLAB 2021a, and the TempCap model is built by Keras 2.0.7 upon Tensorflow 1.2 backend in Python 3.6. The core code corresponding to experiments is publicly available \footnote{\url{https://github.com/sutdqinyan/Multi-stage-SOH}}.

\begin{table*}[!htb]
\caption{Network configurations of the proposed multi-stage SOH estimation model for specific LiBs.}
\renewcommand{\arraystretch}{1.5}
\centering
\scriptsize
\begin{center}
\begin{threeparttable}
\begin{tabular}{c c| c c| c c c c}
\hline
\hline
\multirow{2}{*}{\textbf{Battery}} & \multirow{2}{*}{\textbf{Stage}} & \multicolumn{2}{c|}{\textbf{The proposed approach}} & \multicolumn{2}{c|}{\textbf{LSTM}} & \multicolumn{2}{c}{\textbf{GRU}} \\
\cline{3-8}
& & {Parameters} & {Network Structure} & {Parameters} & \multicolumn{1}{c}{Network Structure} & \multicolumn{1}{|c}{Parameters} & {Network Structure} \\
\hline
\multirow{5}{*}{B7} & Stage 1 & {Epochs=300} & LSTM(30) & {Epochs=100} & LSTM(50) & {Epochs=200} & GRU(15) \\
\cline{2-8}
& \multirow{4}{*}{Stage 2}  & \multirow{4}{*}{Epochs=50} & BCL: Filters=32, Kernel size=[1,2], Strides=[1,2] & \multirow{4}{*}{Epochs=60} & \multirow{4}{*}{LSTM(200)} & \multirow{4}{*}{{Epochs=60}} & \multirow{4}{*}{GRU(30)} \\
&  &  & No. of capsules=8, Dimensions=4 \\
\cline{4-4}
& & & ACL: No. of capsules=4, Dimensions=8 & & & & \\
\cline{4-4}
& & & TRL: LSTM(16) \\
\cline{2-8}
& Stage 3 & {Epochs=500} & LSTM(100)/LSTM(100)/FC(50)  & {Epochs=100} & LSTM(50) & {Epochs=400} & GRU(40) \\
\hline
\multirow{5}{*}{B6} & Stage 1 & {Epochs=300} & LSTM(200) & {Epochs=100} & LSTM(50) & {Epochs=100} & GRU(30)\\
\cline{2-8}
& \multirow{4}{*}{Stage 2} & \multirow{4}{*}{Epochs=50} & BCL: Filters=32, Kernel size=[1,2], Strides=[1,2] & \multirow{4}{*}{Epochs=60} & \multirow{4}{*}{LSTM(200)} & \multirow{4}{*}{Epochs=60} & \multirow{4}{*}{GRU(50)} \\
&  &  & No. of capsules=8, Dimensions=4 \\
\cline{4-4}
& & & ACL: No. of capsule=4, Dimensions=8 & & & & \\
\cline{4-4}
& & & TRL: LSTM(16) \\
\cline{2-8}
& Stage 3 & {Epochs=300} & LSTM(800)/FC(60) & {Epochs=100} & LSTM(40) & {Epochs=150} & GRU(20)\\
\hline
\multirow{5}{*}{B5} & Stage 1 & {Epochs=200} &LSTM(200) & {Epochs=100} & LSTM(200) & {Epochs=100} & GRU(200)\\
\cline{2-8}
& \multirow{4}{*}{Stage 2} & \multirow{4}{*}{Epochs=50} & BCL: Filters=32, Kernel size=[1,2], Strides=[1,2] & \multirow{4}{*}{Epochs=60} & \multirow{4}{*}{LSTM(50)} & \multirow{4}{*}{Epochs=60} & \multirow{4}{*}{GRU(50)} \\
&  &  & No. of capsules=8, Dimensions=4 \\
\cline{4-4}
& & & ACL: No. of capsule=4, Dimensions=8 & & & & \\
\cline{4-4}
& & & TRL: LSTM(16) \\
\cline{2-8}
& Stage 3 & {Epochs=300} & LSTM(300)/FC(200) & {Epochs=60} & LSTM(50) & {Epochs=60} & GRU(50)\\
\hline
\hline
\end{tabular}
\end{threeparttable}
\end{center}
\end{table*}

\begin{table}[!htb]
\caption{Performance comparisons between the proposed multi-stage SOH estimation model with its counterparts using $RMSE (\%)$.}
\renewcommand{\arraystretch}{1.5}
\centering
\scriptsize
\begin{center}
\begin{threeparttable}
\begin{tabular}{c| c| c| c c c}
\hline
\hline
\multicolumn{1}{c|}{\multirow{2}{*}{\textbf{Source Battery}}} & \multicolumn{1}{c|}{\multirow{2}{*}{\textbf{Type}}} & \multicolumn{1}{c|}{\multirow{2}{*}{\textbf{Method}}} & \multicolumn{3}{c}{{$RMSE (\%)$)}}\\
\cline{4-6}
& & & Stage 1 & Stage 2 & Stage 3 \\
\cline{1-6}
\multirow{5}{*}{B7} & \multirow{3}{*}{Multi-stage} & \textbf{Proposed} & \textbf{{0.32$\pm$0.06}} & \textbf{{1.18$\pm$0.09}} & \textbf{{0.42$\pm$0.02}} \\
& & {GRU}   & 0.40$\pm$0.07 & {1.58$\pm$0.12} & 1.98 $\pm$0.15  \\
& & {LSTM} & 0.49$\pm$0.08 & {2.57$\pm$0.17} & 2.91$\pm$0.16  \\
\cline{2-6}
& \multirow{2}{*}{Non-stage} & {GPR-CC [26]}  & \multicolumn{3}{c}{0.78}   \\
& & {LSTM-TL [8]} & \multicolumn{3}{c}{{0.49}} \\
\hline
\multirow{5}{*}{B6} & \multirow{3}{*}{Multi-stage} & \textbf{Proposed} & \textbf{{0.57$\pm$0.04}} & \textbf{{0.89$\pm$0.05}} & \textbf{{0.66$\pm$0.03}} \\
& & {GRU}   & 2.00 $\pm$0.13 & {2.03} $\pm$0.21 & 2.36 $\pm$0.08 \\
& & {LSTM} & 2.30$\pm$0.11 & {2.03}$\pm$0.18 & 5.09 $\pm$0.31  \\
\cline{2-6}
& \multirow{2}{*}{Non-stage} & {GPR-CC [26]}  & \multicolumn{3}{c}{1.49} \\
& & {LSTM-TL [8]} & \multicolumn{3}{c}{Not available} \\
\hline
\multirow{5}{*}{B5} & \multirow{3}{*}{Multi-stage} & \textbf{Proposed} & \textbf{{0.23$\pm$0.02}} & \textbf{{0.38$\pm$0.03}} & \textbf{{0.47$\pm$0.02}} \\
& & {GRU}  & 0.41$\pm$0.05 & {3.46}$\pm$0.26 & 2.02$\pm$0.06  \\
& & {LSTM} & 0.69$\pm$0.07 & {1.39}$\pm$0.11 & 1.58$\pm$0.04  \\
\cline{2-6}
& \multirow{2}{*}{Non-stage} & {GPR-CC [26]} & \multicolumn{3}{c}{0.95} \\
& & {LSTM-TL [8]} & \multicolumn{3}{c}{Not available} \\
\hline
\hline
\end{tabular}
\end{threeparttable}
\end{center}
\end{table}

\subsection{Data Reconstruction in Phase Subspace}
This section manipulates the raw cycling data with phase space, through which insightful dynamics are revealed with the high-dimensional representation. Two crucial parameters in Eq. (5), i.e., the time delay $\tau$ and dimension $r$, are specified as 3 and 5 for all cycles, respectively. In this way, the original one-dimensional signal, including discharge voltage, current, and temperature, is extended into a three-dimensional space. Fig. 5 visualizes the cycling data from the first stage of B7 before and after data reconstruction. Compared to the original one-dimensional time series given in Fig. 5(a), the differences over cycles are apparent for the reconstructed data, as shown in Fig. 5(b), specifically for the discharge voltage and temperature influenced by the performance degradation. It makes sense to infer that the cycle discrepancy can be amplified with PSR, and the total dimension space is nine for each discharging cycle.

\subsection{Cycle Discrepancy Learning}
After data reconstruction, cycling consistency components and corresponding discrepancy components over cycles are decomposed according to procedures in Section III.B. Through attempting the tunable parameter $S$ in Eq. (8), four consistent and the remaining five inconsistent features over cycles in each stage are decomposed from B7, as shown in Figs. 6 and 7. The blue and red lines indicate 70$\%$ training cycles and 30$\%$ testing cycles, respectively. Several observations are summarized:

\begin{itemize}
  \item As shown in Fig. 6, consistent features keep the similar behavior in the training and testing cycles, demonstrating excellent generation ability. The meaning of each consistent feature can be inferred in combination with the original discharge voltage, current, and temperature signals. Specifically, the first feature is almost stationary, which may be the average time series of the discharge signals. The second and fourth features are decreasing, which may have a close relationship with the voltage. The trend of the third component increases and may have a close relationship with the temperature.
  \item Fig. 7 shows the discrepancy features in all stages, varying from cycle to cycle for both training and testing cycles. The performance degradation is driven by these discrepancy features, and the discrepancy becomes significant with the increase of the cycle.
\end{itemize}

\subsection{The Transferable Multiple-stage Source Model}
The source models are developed using cycling discrepancy components from one battery, and then the transferability is verified on other batteries. For a fair comparison with existing approaches \cite{2020TIETrasnferLSTM} \cite{YANG2018387}, battery B7 is utilized to build the source model, and subsequently more comparisons are provided.

\begin{figure*}[!ht]
	\centering
	\subfigure[B7 as source model in Stage 1]
	{
	\begin{minipage}[t]{0.3\linewidth}
	\centering
	\includegraphics[width=5cm]{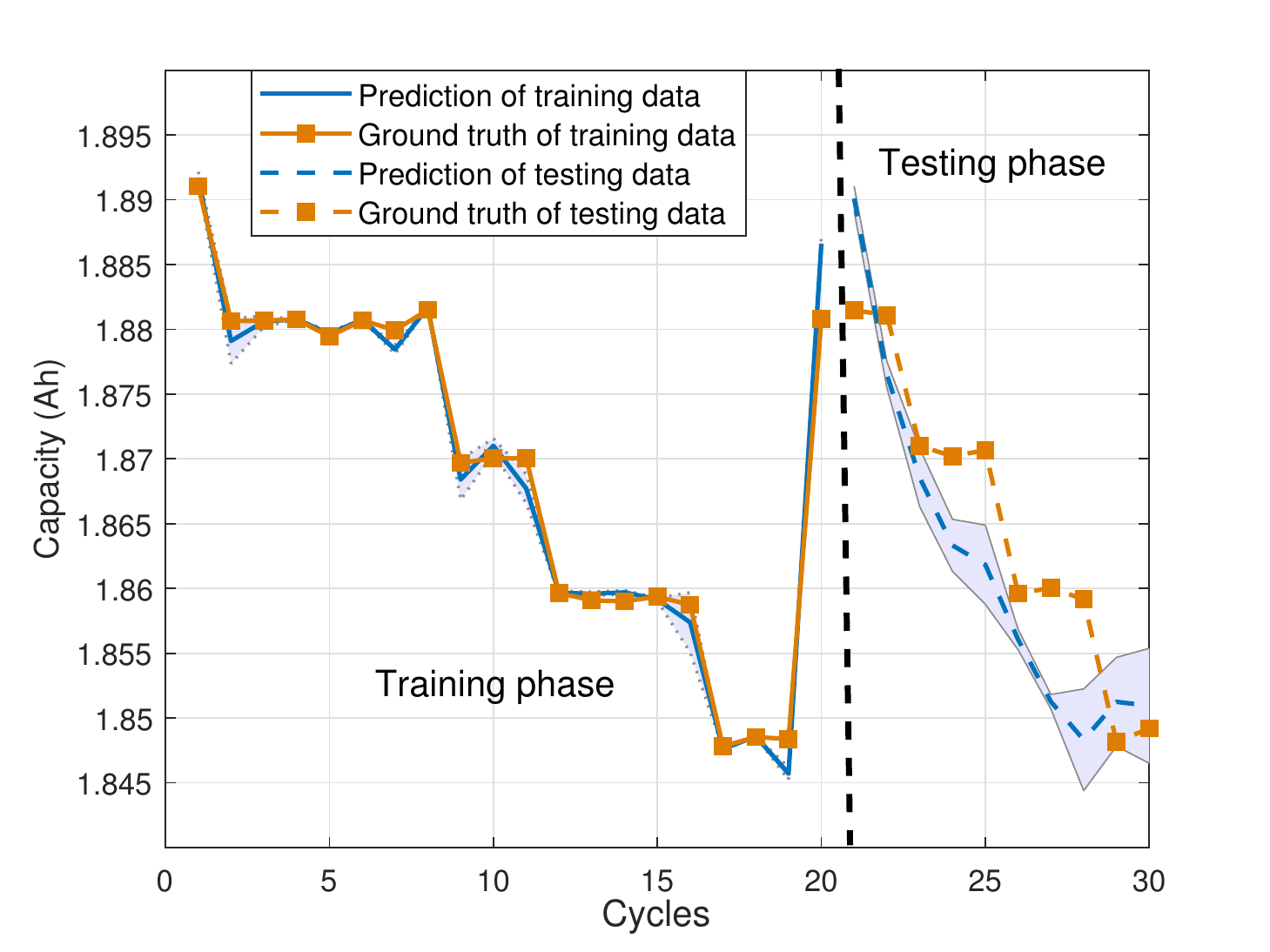}
	\end{minipage}
	}
	\subfigure[B6 as source model in Stage 1]
	{
	\begin{minipage}[t]{0.3\linewidth}
	\centering
	\includegraphics[width=5cm]{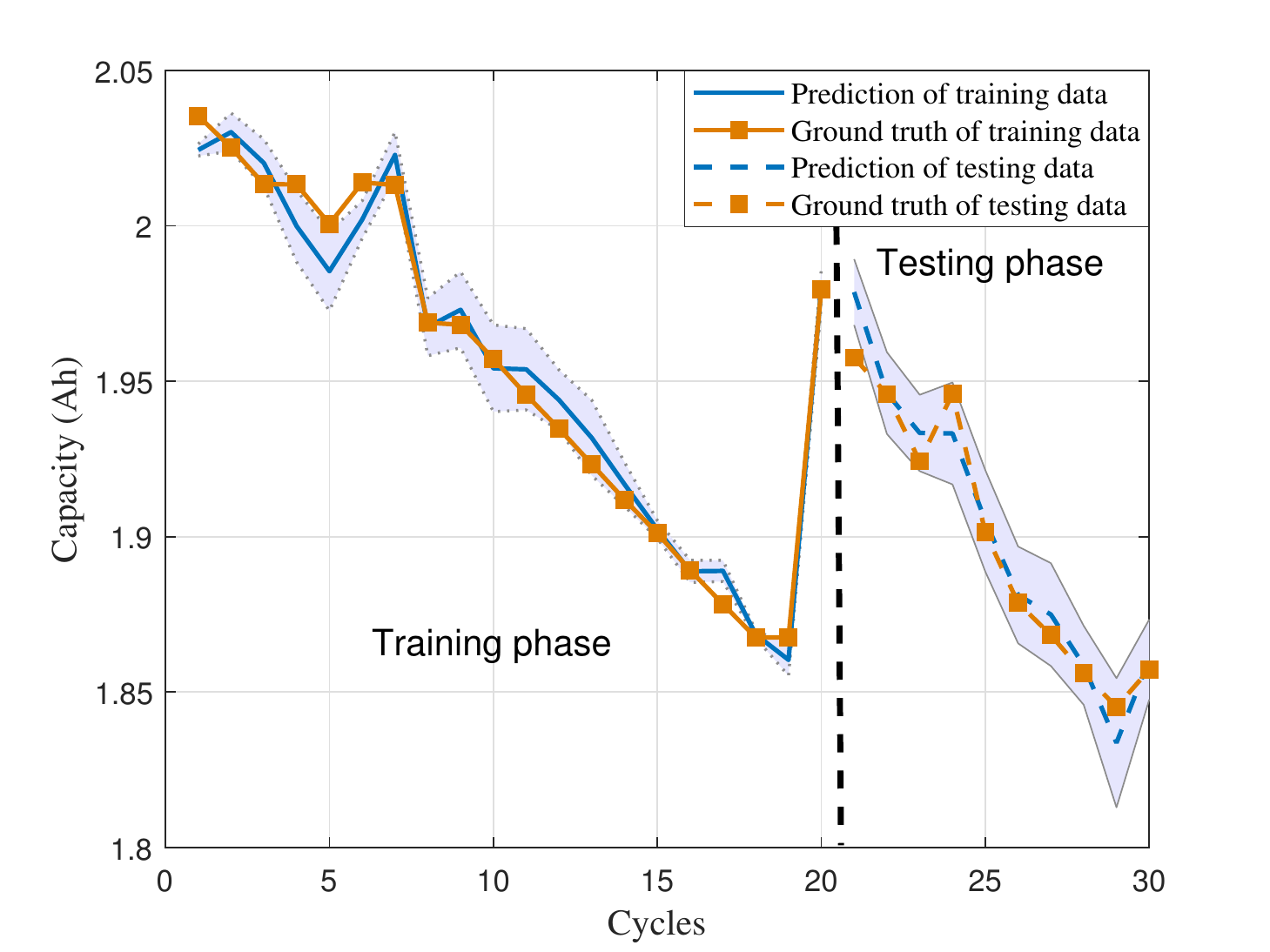}
	\end{minipage}
	}
	\subfigure[B5 as source model in Stage 1]
	{
	\begin{minipage}[t]{0.3\linewidth}
	\centering
	\includegraphics[width=5cm]{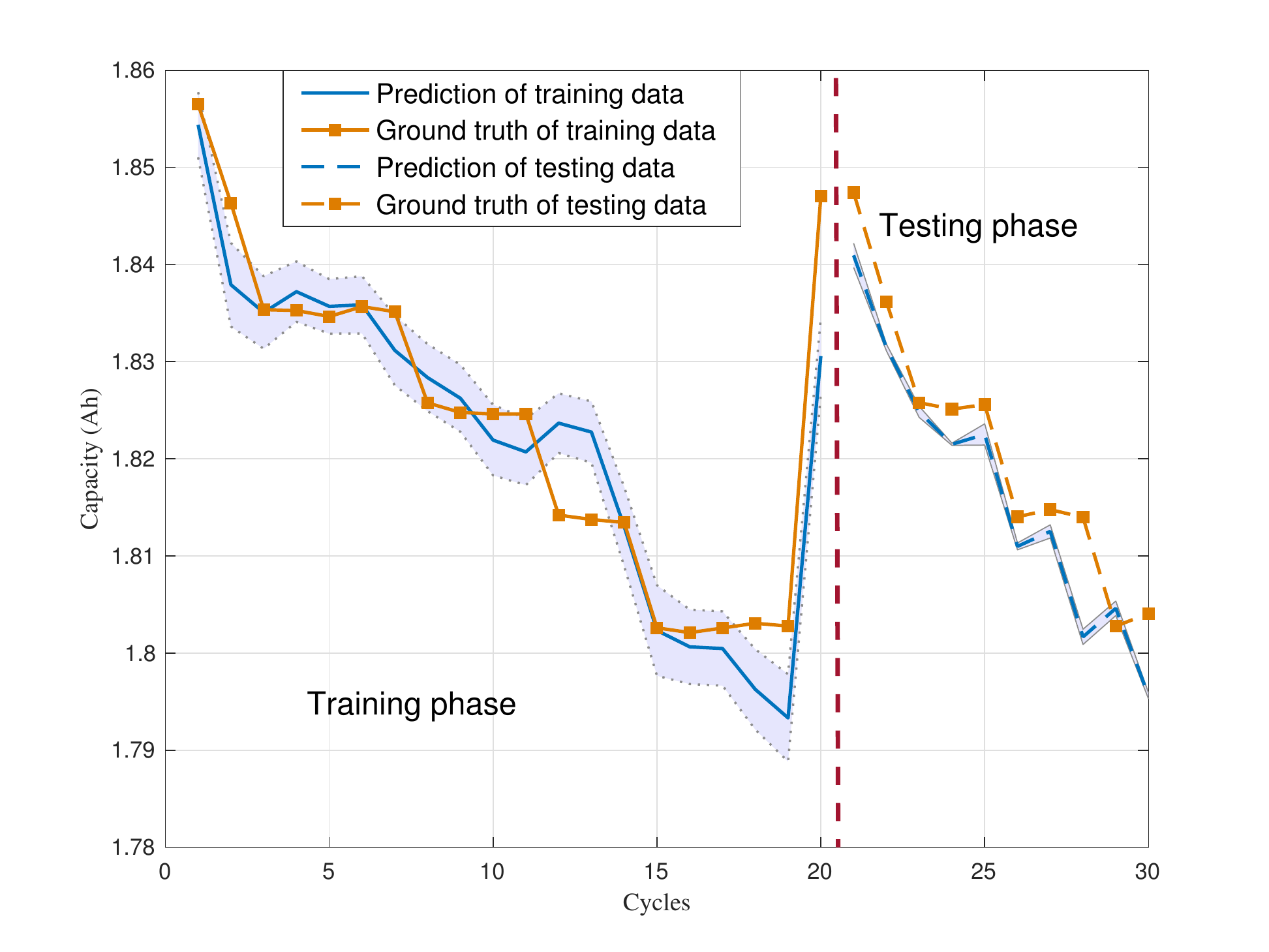}
	\end{minipage}
	}
	
	\subfigure[B7 as source model in Stage 2]
	{
	\begin{minipage}[t]{0.3\linewidth}
	\centering
	\includegraphics[width=5cm]{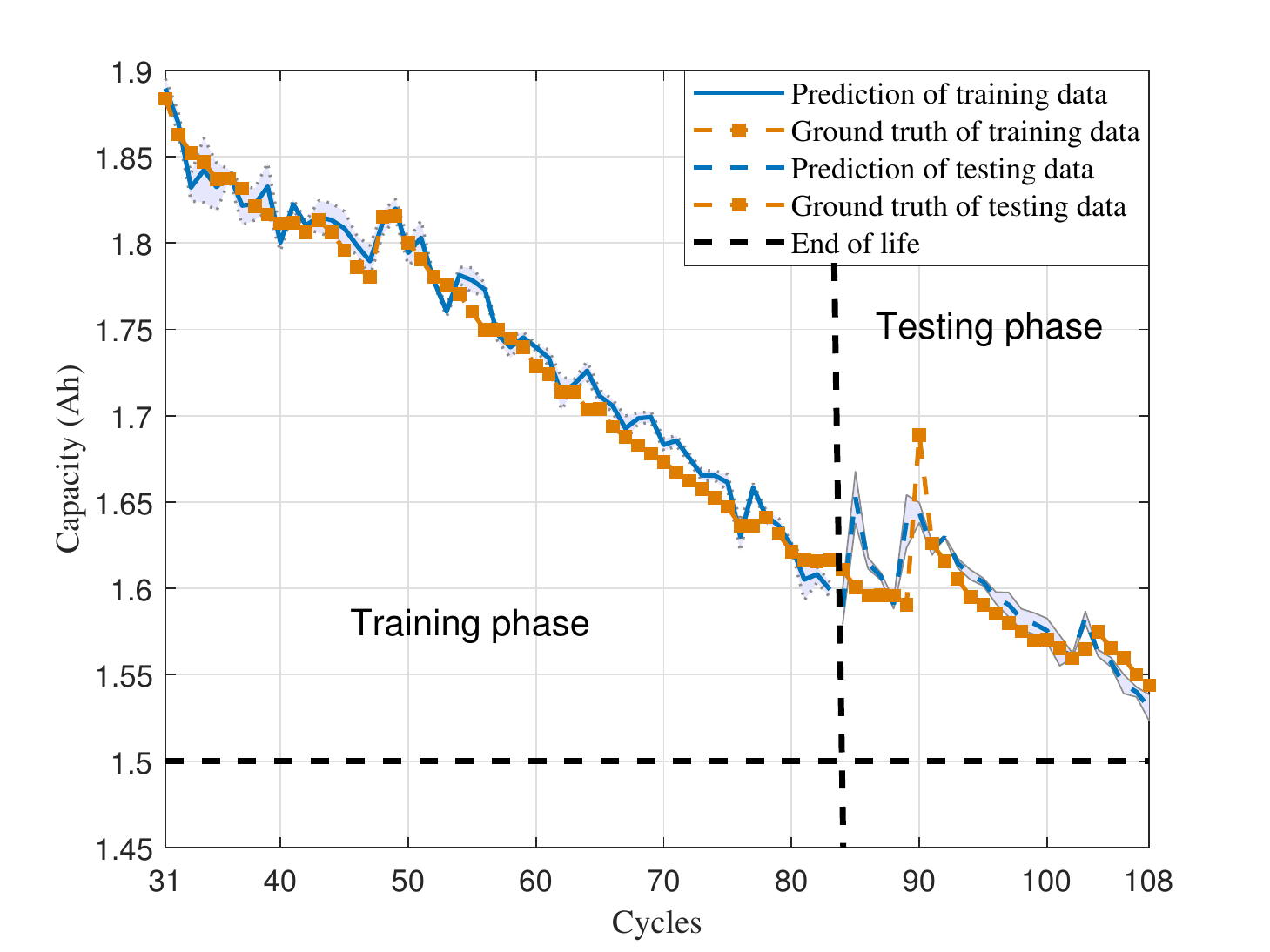}
	\end{minipage}
	}
	\subfigure[B6 as source model in Stage 2]
	{
	\begin{minipage}[t]{0.3\linewidth}
	\centering
	\includegraphics[width=5cm]{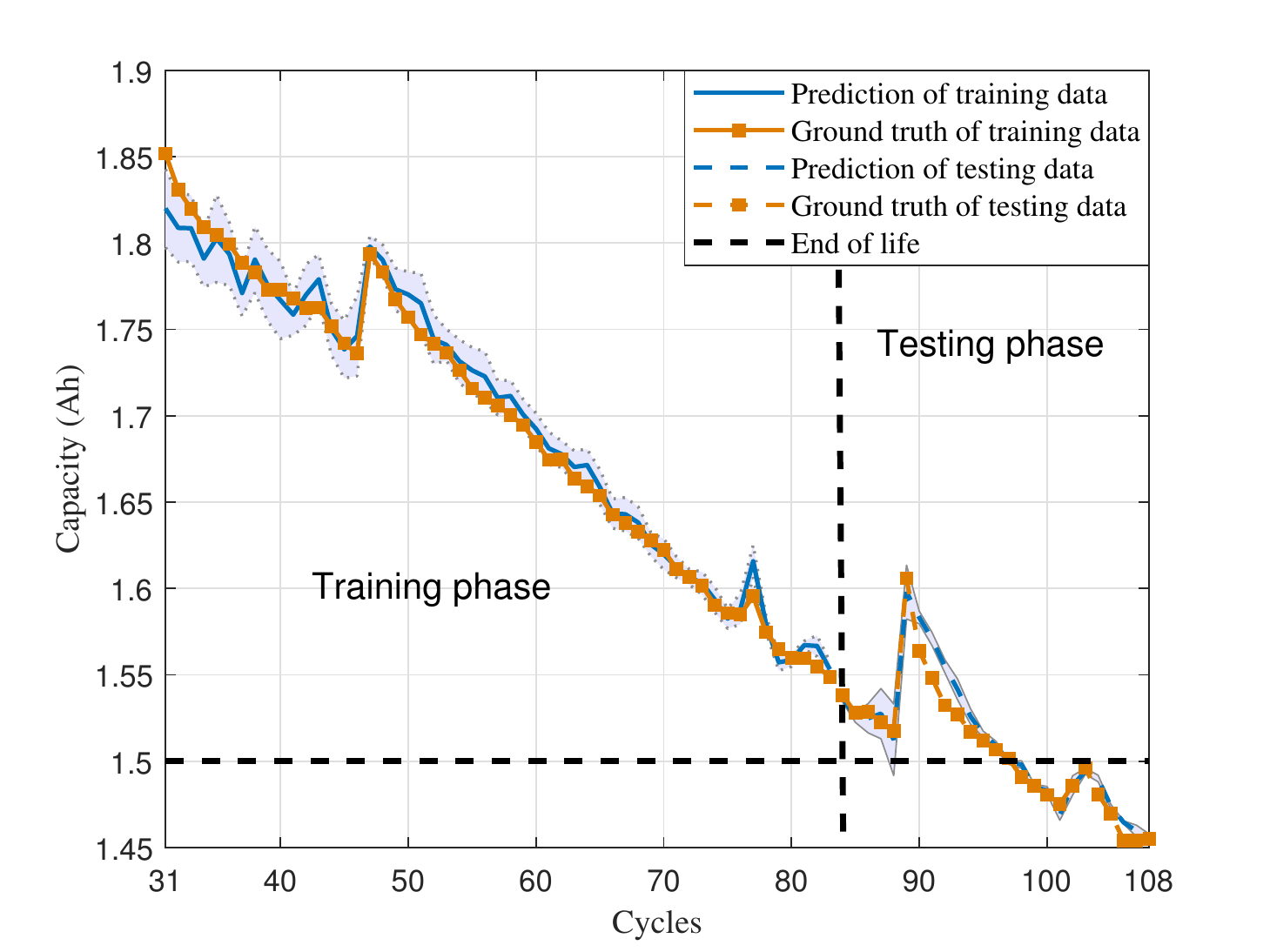}
	\end{minipage}
	}
	\subfigure[B5 as source model in Stage 2]
	{
	\begin{minipage}[t]{0.3\linewidth}
	\centering
	\includegraphics[width=5cm]{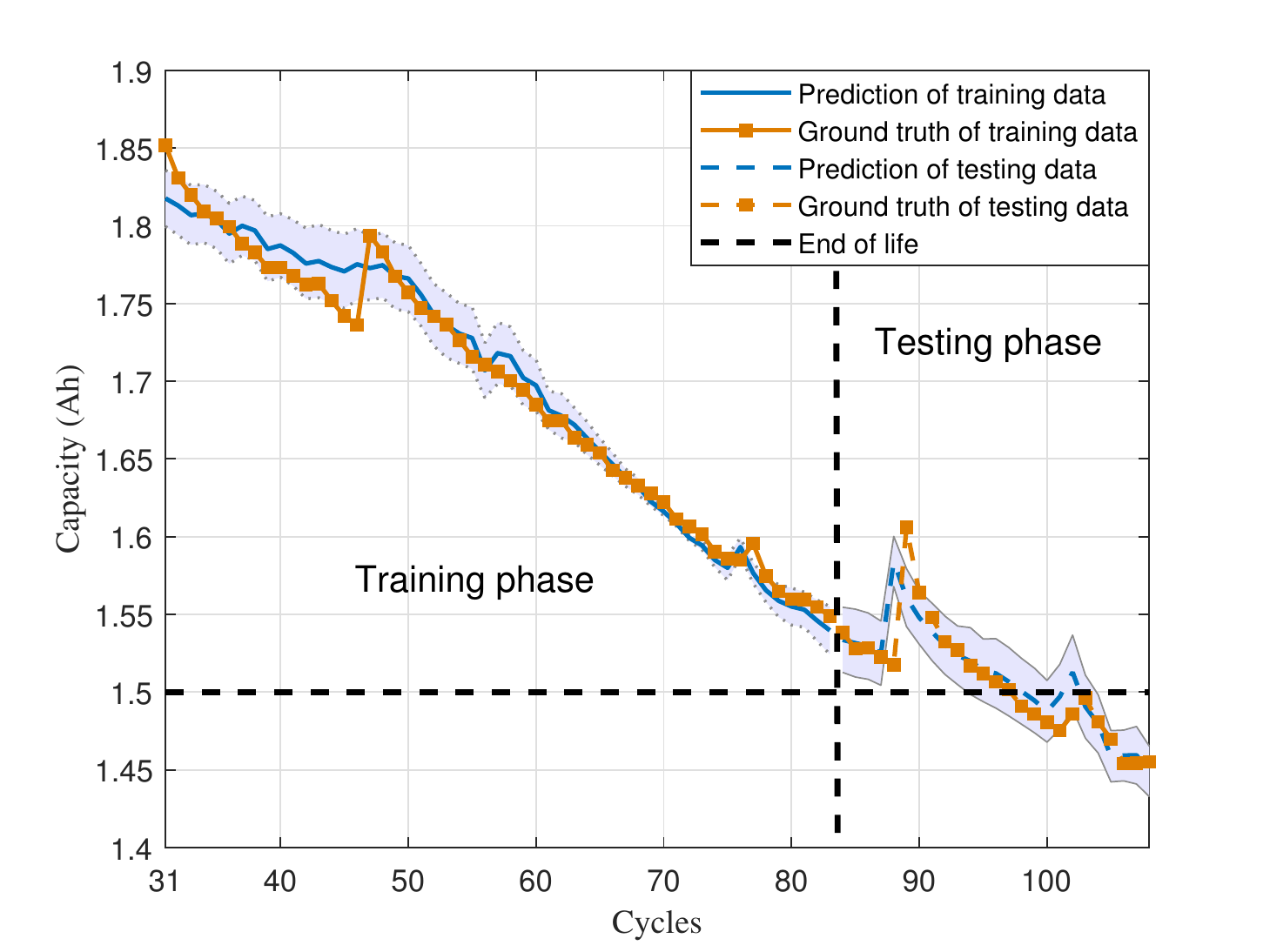}
	\end{minipage}
	}
	
	\subfigure[B7 as source model in Stage 3]
	{
	\begin{minipage}[t]{0.3\linewidth}
	\centering
	\includegraphics[width=5cm]{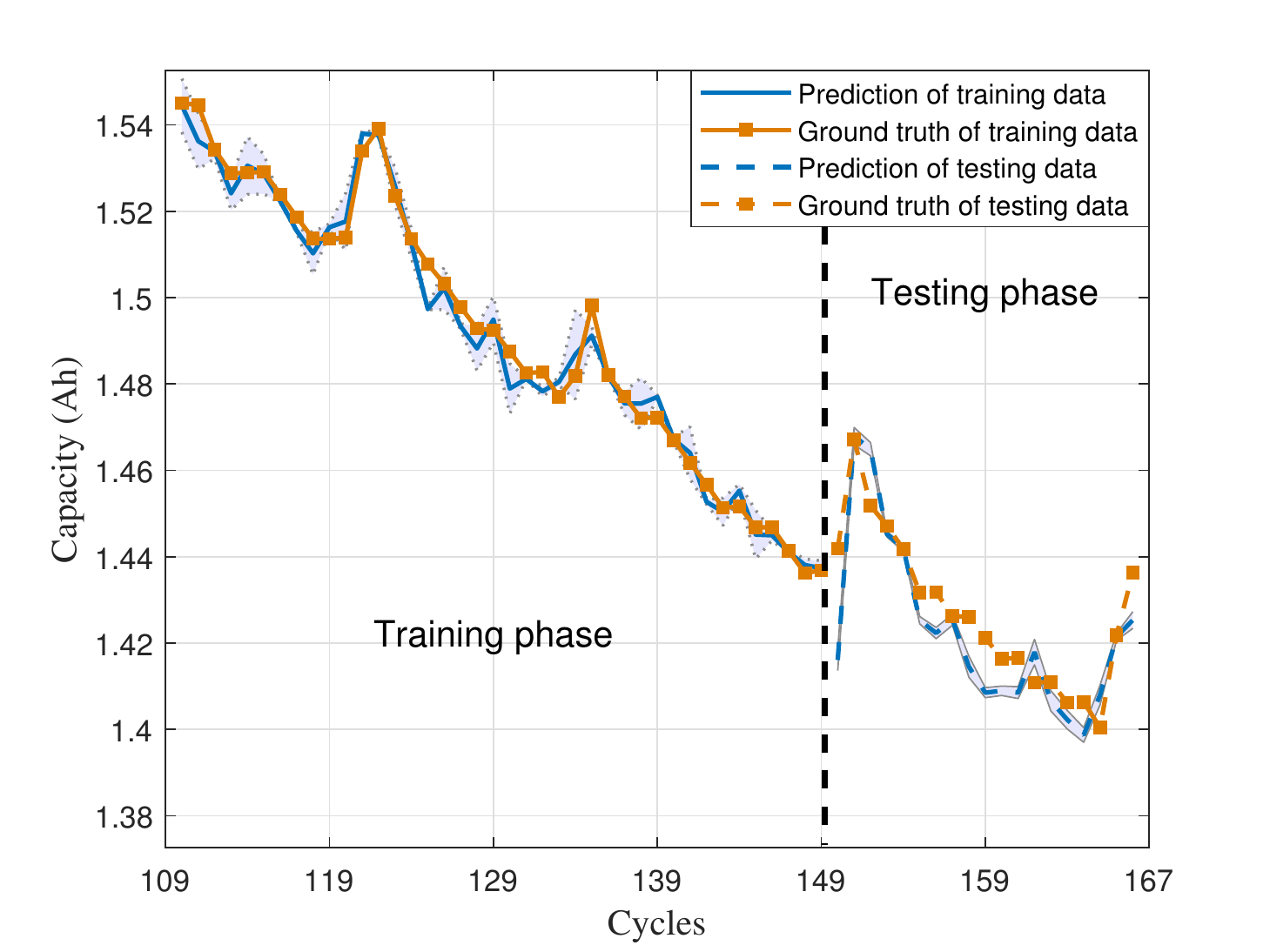}
	\end{minipage}
	}
	\subfigure[B6 as source model in Stage 3]
	{
	\begin{minipage}[t]{0.3\linewidth}
	\centering
	\includegraphics[width=5cm]{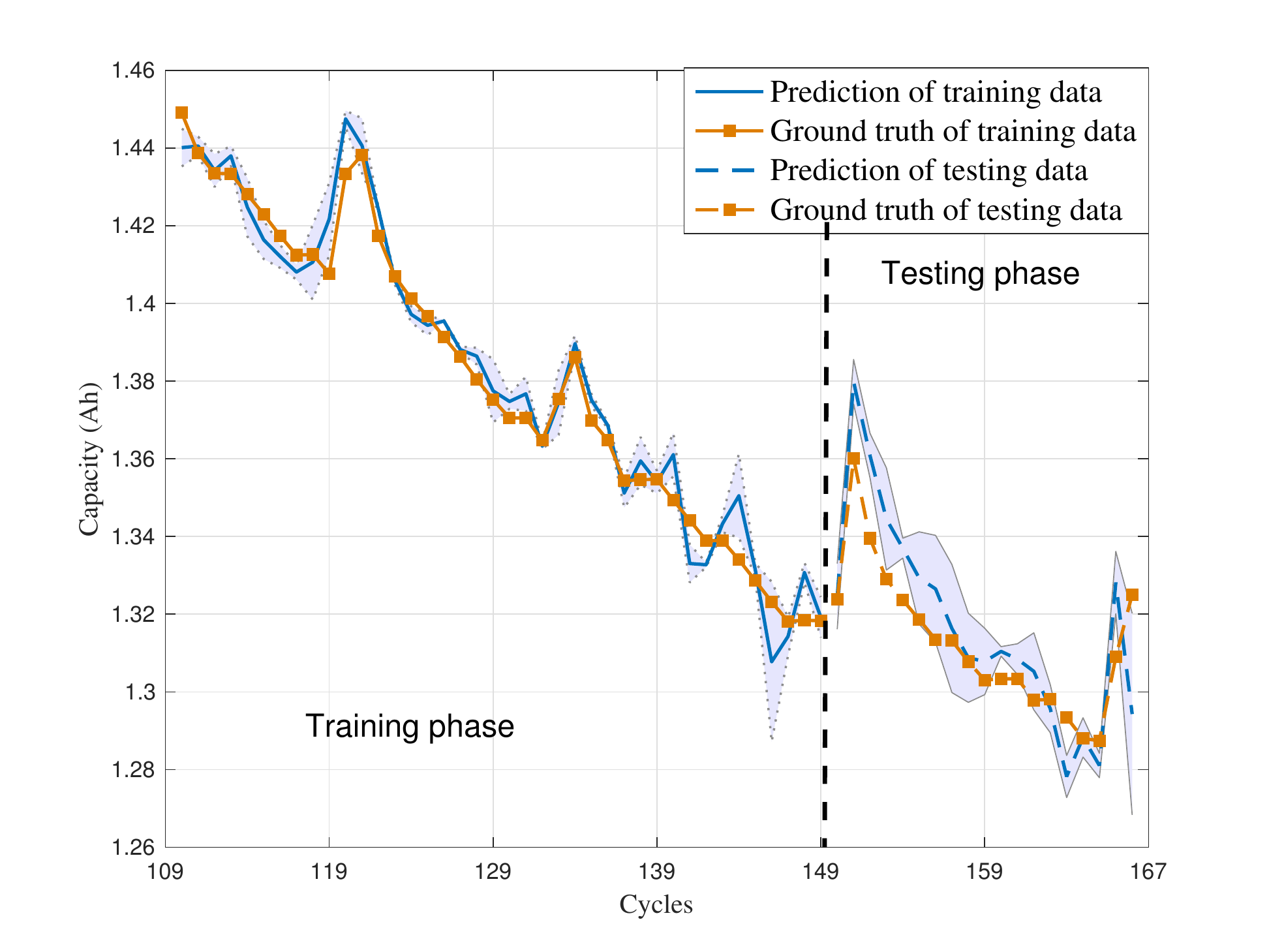}
	\end{minipage}
	}	
	\subfigure[B5 as source model in Stage 3]
	{
	\begin{minipage}[t]{0.3\linewidth}
	\centering
	\includegraphics[width=5cm]{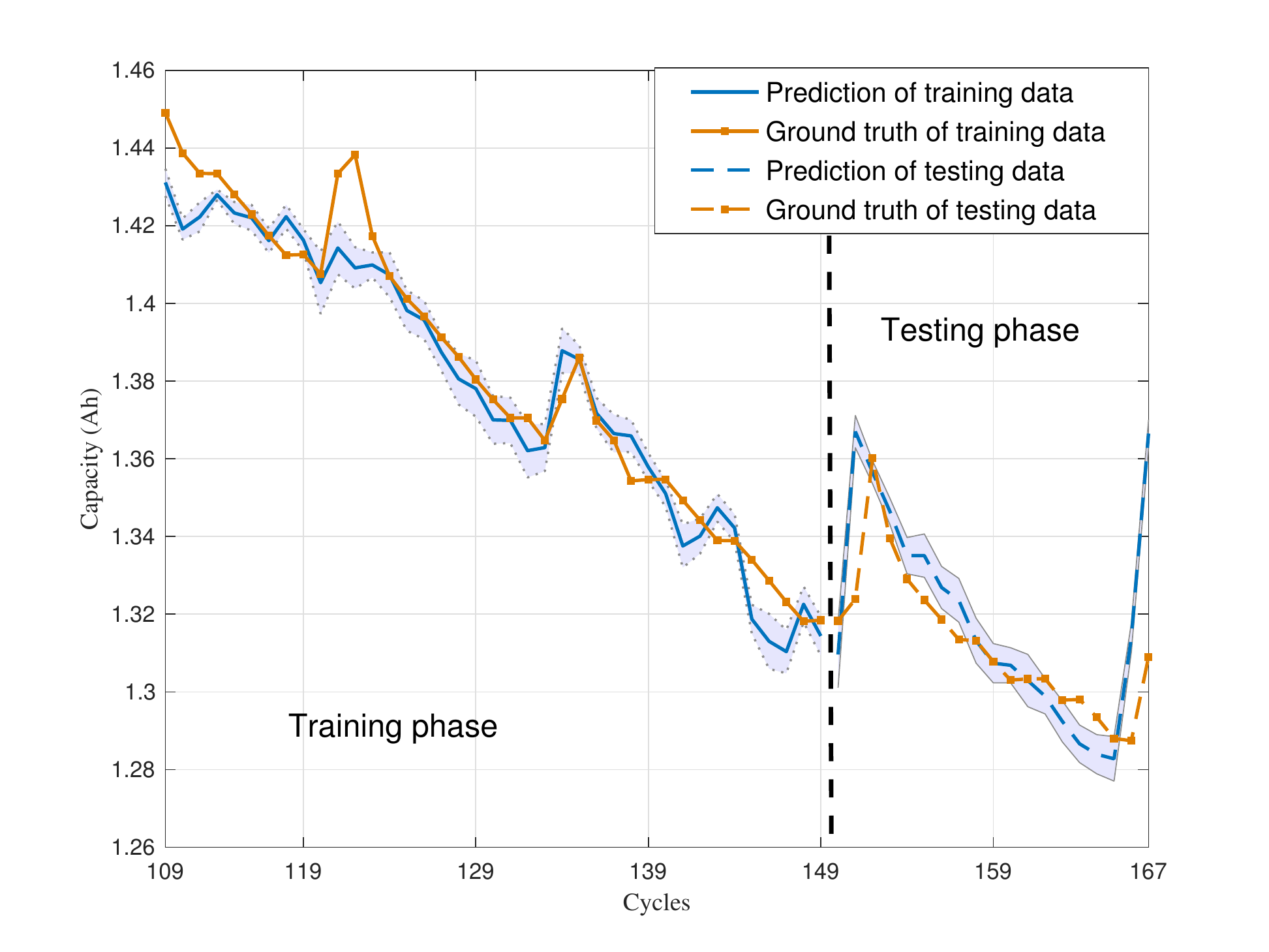}
	\end{minipage}
	}
	\caption{The estimation performance of training phase and testing phase when batteries B7, B6, and B5 are adopted as source models in each stage.}
	\label{Fig8}
\end{figure*}

\subsubsection{Source Model for Stages using LSTM}
Since training cycles of B7 in Stages 1 and 3 are less than 50, a lightweight LSTM is preferred as the backbone to develop source models. The first 70$\%$ cycles of each stage are used as training data, i.e., the first 20 cycles in Stage 1 and the first 40 cycles in Stage 3. Estimation results are tested with the remaining cycles with the well-trained model.

With the network configuration listed in Table II, the specific estimation results for each stage of B7 are plotted in Figs. 8(a), 8(d), and 8(g), including training and testing phases. By looking into the results, we have the following understandings:
\begin{itemize}
\item The ground truth of battery capacity is non-stationary, decreasing with the increase of cycles. The predictions of the proposed method are capable of capturing the degradation trend very well in the testing phase, signifying the good generalization ability of the proposed method. Moreover, the regeneration phenomenon of battery capacity can be well predicted.
\item The proposed method runs ten times to evaluate the average performance and robustness. The generated estimation is close to its ground truth with a narrow variation band (95$\%$ confidence level).
\end{itemize}

Using the index $RMSE$, traditional LSTM and gated recurrent unit (GRU) \cite{2014GRU} are adopted for quantitative comparison with the same stage division information. The network configurations of LSTM and GRU are listed in Table II for clarity, which both are lightweight models.

\begin{table*}[!htb]
\caption{Comparison of various methods by adopting B7 as the source data using index $RMSE (\%)$.}
\renewcommand{\arraystretch}{1.5}
\centering
\scriptsize
\begin{center}
\begin{threeparttable}
\begin{tabular}{c| c| c| c| c c c }
\hline
\hline
\multicolumn{1}{c|}{\multirow{2}{*}{\textbf{Target Data  $\mathfrak{D}_T$}}} & \multicolumn{1}{c|}{\multirow{2}{*}{\textbf{Source Data $\mathfrak{D}_S$}}} & \multicolumn{1}{c|}{\multirow{2}{*}{\textbf{Model Type}}}  & \multicolumn{1}{c|}{\multirow{2}{*}{\textbf{Method}}} & \multicolumn{3}{c}{${RMSE (\%)}$} \\
\cline{5-7}
& & & & Stage 1 & Stage 2 & Stage 3 \\
\cline{1-7}
\multirow{7}{*}{B5} & \multirow{7}{*}{B7} & \multirow{2}{*}{Multi-stage} & \textbf{Proposed} & {\textbf{{0.39}}} &  \textbf{{0.50}} & 0.62 \\
& & & {IMA} & {27.78\%} & {7.41\%} & {-14.81\%} \\
\cline{3-7}
& & \multirow{6}{*}{Non-stage} & {LSTM-TL \cite{2020TIETrasnferLSTM}} & \multicolumn{3}{c}{0.54} \\
& & & {CADA \cite{2020CADA}} & \multicolumn{3}{c}{1.97} \\
& & & {DNN \cite{2020TIETrasnferLSTM}} & \multicolumn{3}{c}{2.58} \\
& & & {GPR \cite{2020TIETrasnferLSTM}} & \multicolumn{3}{c}{1.39} \\
& & & {SVR \cite{WEI2018352}} & \multicolumn{3}{c}{7.97} \\
& & & {GMDH \cite{HU2012359}} & \multicolumn{3}{c}{3.27} \\
\cline{1-7}
\multirow{7}{*}{B6} & \multirow{7}{*}{B7} & \multirow{2}{*}{Multi-stage} & \textbf{Proposed} & {\textbf{{0.85}}} & \textbf{{0.98}} & \textbf{{0.82}} \\
& & & {IMA} & {40.56\%} & {31.47\%} & {42.66\%} \\
\cline{3-7}
& & \multirow{5}{*}{Non-stage} & {LSTM-TL \cite{2020TIETrasnferLSTM}} & \multicolumn{3}{c}{1.43} \\
& & & {CADA \cite{2020CADA}} & \multicolumn{3}{c}{2.14} \\
& & & {DNN \cite{2020TIETrasnferLSTM}} & \multicolumn{3}{c}{2.25} \\
& & & {GPR \cite{2020TIETrasnferLSTM}} & \multicolumn{3}{c}{1.66} \\
& & & {SVR \cite{WEI2018352}} & \multicolumn{3}{c}{2.13} \\
& & & {GMDH \cite{HU2012359}} & \multicolumn{3}{c}{11.14} \\
\hline
\hline
\end{tabular}
\end{threeparttable}
\end{center}
\begin{tablenotes}
\item[1] $^{[1]}$ IMA is short for the improved accuracy compared to the results of LSTM-TL.
\end{tablenotes}
\end{table*}

\begin{table*}[!ht]
\centering
\caption{Performance of the proposed method by adopting various source models using index $RMSE (\%)$.}
\scriptsize
\renewcommand{\arraystretch}{1.5}
\begin{center}
\begin{threeparttable}
\begin{tabular}{c|c|c|c|c|c|c}
\hline
\hline
\multirow{2}{*}{\textbf{Target Data $\mathfrak{D}_T$}} & \multirow{2}{*}{\textbf{Source Data $\mathfrak{D}_S$}} & \multicolumn{1}{c|}{\multirow{2}{*}{\textbf{CADA \cite{2020CADA}}}} & \multicolumn{1}{c|}{\multirow{2}{*}{\textbf{LSTM-TL \cite{2020TIETrasnferLSTM}}}} & \multicolumn{3}{c}{\textbf{The proposed method}}  \\
\cline{5-7}
& & & & {{Stage 1}} & {{Stage 2}} & {{Stage 3}}   \\
\hline
\multirow{3}{*} {B6} & {B5} & 2.42 & 2.02 & 0.53 & 1.87 & 0.72  \\
& {B7} & 1.76 & 1.07 & 0.85 & 0.98 & 0.92         \\
& {B5 + B7} & 1.89 & 1.33 & 0.76 & 0.86 & 0.77 \\
\hline
\multirow{3}{*}{B5} & {B6} & 1.54 & 1.32 & 0.39 & 1.16 & 0.99 \\
& {B7} & 1.09 & 0.79 & 0.39 & 0.50 & 0.62         \\
& {B6 + B7} & 1.11 & 0.88 & 0.18 & 0.81 & 0.54 \\
\hline
\multirow{3}{*}{B7} & {B5} & 1.87 & 1.45 & 0.57 & 1.14 & 0.71 \\
& {B6} & 1.34 & 1.23 & 0.49 & 0.93 & 0.95         \\
& {B5 + B6} & 1.26 & 1.21 & 0.36 & 0.91 & 0.91 \\
\hline
\hline
\end{tabular}
\end{threeparttable}
\end{center}
\end{table*}

Contributing to finding the degradation-related information with the help of phase space reconstruction and cycling discrepancy components, the estimation errors of the proposed method are below $0.5 \%$ for B7 in both Stages 1 and 2. For a fair comparison, the counterparts, including GRU and LSTM, are repetitively run ten times to evaluate the average performance with a standard variation. Overall, the proposed method outperforms both GRU and LSTM in Stages 1 and 3 regarding accuracy and robustness, as shown in Table III. Specifically, in Stage 3, the estimation accuracy of GRU and LSTM decrease significantly. Furthermore, two non-stage models treating the entire degradation process in a single behavior are employed for comparison, i.e., LSTM-TL \cite{2020TIETrasnferLSTM} and GPR-CC \cite{YANG2018387}. Both models take the first 70$\%$ cycles of the whole process as training data and the remaining ones as testing data. As the testing cycles mainly belong to the divided Stage 3, it makes sense to compare the estimation results in Stage 3 of the proposed method with that of LSTM-TL \cite{2020TIETrasnferLSTM} and GPR-CC \cite{YANG2018387}. As such, the proposed method still outperforms both counterparts with the training data only from Stage 3. It is more reasonable to infer future capacity with historical cycling data belonging to the same stage. We have observed the same conclusions for B6 and B5. Specifically, the estimation results in Stage 3 of the proposed method are much less than that of GPR-CC \cite{YANG2018387}. Detailed estimation results for B5 and B6 are visualized in Figs. 8(b), 8(h), 8(c), and 8(i).

Since the network structure of LSTM and TemCap is not complex, empirical knowledge is used to configure the hyper-parameter \cite{2021TIISOC_QIN} \cite{LSTMnode}, and good estimation results are obtained without many tuning burdens. Alternatively, automatic hyper-parameter tuning is regarded as a promising solution, such as neural architecture search \cite{LiuPara1}, Bayesian optimization \cite{LiuPara2}.

\subsubsection{Source Model for Stages using TemCap}
The number of training data of Stage 2 is more than 50 cycles. The designed TemCap is adopted for Stage 2 to capture the deep features, following the same data preparation with the other two stages.

According to the network structure of Stage 2 specified in Table II, the specific estimation results of the proposed method for B7, B6, and B5 are plotted in Figs. 8(d), 8(e), and 8(f), respectively. The tracking trend of predictions meets well with the ground truth. Moreover, for B6 and B5, predictions reach the threshold of the end of life at the actual cycle in the testing phase. As such, degraded batteries can be timely replaced without any delay. Following the network configurations given in Table II, the average estimation results of GRU and LSTM for B7 are 1.58 and 2.57, which are larger than the 1.18 yielded by the proposed method. Besides, their estimation variations are also larger than that of the proposed method. As mentioned earlier, the testing data of LSTM-TL \cite{2020TIETrasnferLSTM} and GPR-CC \cite{YANG2018387} are mainly distributed in Stage 3. Therefore, a large estimation error of the proposed method in Stage 2 doesnot mean a performance reduction. On the contrary, the proposed method yields even better results for B6 and B5 in Stage 2. The detailed comparisons between the proposed method and its counterparts are summarized in Table III, in which the results marked in blue indicate that the proposed method yields better performance.

\subsection{Transfer Model for Target Domain}
With the developed source model of each battery, this part verifies the transferability of the proposed method according to the details given in Section III.D.

For a fair comparison with existing methods, we first employ battery B7 as the source model and attempt to verify the estimation errors under TL for other batteries. By checking the consistency features of B5 with B7, they follow the same statistical pattern, which indicates that source models of B7 can be directly applied for the target task with B5. Two representative algorithms are implemented for comparison given as follows:
\begin{itemize}
\item Parameter transfer-based LSTM (LSTM-TL) \cite{2020TIETrasnferLSTM}: LSTM-TL gains transferability by inheriting LSTM layers from source data and newly updated fully connected layers with the target data.
\item Contrastive adversarial domain adaptation (CADA) \cite{2020CADA}: CADA employs a contrastive loss to consider target-specific information when learning domain invariant features. The same LSTM configuration listed in Table II has been adopted as the source and target networks.
\end{itemize}

As listed in Table IV, the estimation errors of the proposed method obtained for Stages 1 and 2 are 0.39$\%$ and 0.50$\%$, respectively, yielding the best results compared to current algorithms. Although the estimation error in Stage 3 is slightly higher than that of LSTM-TL, it is worth noting that the source model developed by B7 is directly applied for B5 without fine-tuning the source model like LSTM-TL. For CADA, it is observed that estimation errors are large compared with the proposed method. This may be caused by the strong non-stationarity of the battery degradation data. In contrast, the good prediction ability of the proposed method may largely be attributed to the extracted domain-invariant features, which also can explain the difference across batteries.

The consistent components between B7 and B6 are different, which may be caused by the slight manufacturing difference. As such, a shallow neural network with 50 neurons is constructed to compensate for the estimation error of the source model in each stage within the first ten cycles. According to the updating strategy given in Section III.D, we obtain the initial estimation results of the source model, and the final estimation is updated by adding the compensation. After that, the transfer estimation errors of B6, as listed in Table IV, are 0.85\%, 0.98\%, and 0.82\% in each stage, respectively. The prediction accuracy of the proposed method is much smaller than that of LSTM-TL and CADA.

To verify the transferability with more scenarios, Table V lists extensive comparisons by employing each battery as the target task. Correspondingly, batteries B5, B7, and their mixture served as the source tasks in turn. For the target task B6, estimation errors of the baseline model are 0.57\%, 0.89\%, and 0.66\%. We found similar and even better estimation performance than the baseline model by transferring the estimation model developed with other source tasks, like B5, B7. As most estimation errors are less than 1$\%$, it concludes that the proposed method provides an accurate estimation with the transferable multi-stage SOH estimation model.

%

\section{Conclusion and Future Work}
This article puts forward a transferable Lithium-ion battery state-of-health estimation model along with the analysis of multi-stage degradation behavior for the first time. The introduction of multi-stage degradation behavior separates the whole discharging into several stages, each of which holds the unique degradation behavior. To pursue an efficient transferability between source and target tasks, a series of contributions are proved in this study. Phase space reconstruction contributes to revealing the hidden dynamics from limited measurements. Domain-invariant feature decomposition clearly locates the features relevant to capacity degradation. Moreover, a switching strategy has been proposed to update source model development, utilizing long short-term memory network and capsule network. A series of comparisons conducted on a well-known battery benchmark proves that the proposed method has more flexible transferability and accurate estimation compared to its competing approaches. Besides, the following suggestions are recommended as future works:
\begin{itemize}
\item With the capacity fading, the variable data structure with the uneven length is observed among cycling data. The inappropriate disposal of variable data structure leads to information loss or distortion, undermining the estimation accuracy. It is worth finding a proper embedding, like the dynamic time warping, to align the variable data structure for offline training and online application.
\item Instead of assuming three stages, generating stage division results with concurrent consideration of degradation behavior and capacity fading is still an open issue and will generalize the more general transfer learning between LiBs with a different number of degradation stages.
\end{itemize}


\ifCLASSOPTIONcaptionsoff
  \newpage
\fi
\bibliographystyle{IEEEtran}

\end{document}